\documentclass[twoside]{article}

\usepackage[accepted]{aistats2020}

\usepackage[utf8]{inputenc} %
\usepackage[T1]{fontenc}    %
\usepackage{url}            %
\usepackage{booktabs}       %
\usepackage{nicefrac}       %
\usepackage{microtype}      %

\usepackage{hyperref}
\hypersetup{
    colorlinks=true,
    linkcolor=blue,
    urlcolor=blue,
    citecolor=blue
}

\usepackage{dsfont}
\usepackage{bbm}
\usepackage{amsmath, amsthm, amssymb, amsfonts}
\usepackage{bm}
\usepackage{mathtools}
\usepackage{thmtools}
\usepackage[round]{natbib}

\usepackage{xcolor}
\usepackage{booktabs}
\usepackage{multirow}
\usepackage{graphicx}
\usepackage{enumerate}
\usepackage[capitalise,noabbrev]{cleveref}  %
\crefformat{equation}{(#2#1#3)}

\newtheorem{theorem}{Theorem}[section]

\newtheorem{corollary}{Corollary}[section]
\newtheorem{assumption}{Assumption}[section]

\newcommand{\E}{\mathop{\mathbb{E}}}
\newcommand{\V}{\mathop{\mathbb{V}}}
\newcommand{\C}{\mathop{\mathbb{C}}}
\newcommand{\R}{\mathbb{R}}
\newcommand{\tran}{^{\top}}

\DeclareMathOperator{\diag}{diag}
\DeclareMathOperator{\trace}{Tr}

\DeclareMathOperator{\vect}{vec}
\DeclareMathOperator{\sigmoid}{sigmoid}
\DeclareMathOperator{\swish}{swish}
\DeclareMathOperator{\ReLU}{ReLU}
\DeclareMathOperator{\I}{I}

\DeclareMathOperator{\cov}{cov}
\DeclareMathOperator{\var}{var}
\newcommand{\1}{\mathds{1}}
\newcommand{\Dt}{\Delta t}
\newcommand{\xdt}{\boldsymbol{x}}

\let\norm\undefined
\DeclarePairedDelimiter\norm{\lVert}{\rVert}
\DeclarePairedDelimiterX{\dprod}[2]{\langle}{\rangle}{#1, #2}

\begin{document}

\twocolumn[

\aistatstitle{Infinitely deep neural networks as diffusion processes}

\aistatsauthor{
    Stefano Peluchetti\\
    \texttt{speluchetti@cogent.co.jp}\\
    \And
    Stefano Favaro\\
    \texttt{stefano.favaro@unito.it}\\
}

\aistatsaddress{
    Cogent Labs\\
    \And
    Department ESOMAS\\ University of Torino\\ and Collegio Carlo Alberto\\
} ]

\begin{abstract}
When the parameters are independently and identically distributed (initialized) neural networks exhibit undesirable properties that emerge as the number of layers increases, e.g. a vanishing dependency on the input and a concentration on restrictive families of functions including constant functions. We consider parameter distributions that shrink as the number of layers increases in order to recover well-behaved stochastic processes in the limit of infinite depth. This leads to set forth a link between infinitely deep residual networks and solutions to stochastic differential equations, i.e. diffusion processes. We show that these limiting processes do not suffer from the aforementioned issues and investigate their properties.
\end{abstract}

\section{Introduction}\label{sec:introduction}

Modern neural networks (NN) models featuring a large number of layers (depth) and features per layer (width) have achieved a remarkable performance across many domains \citep{lecun2015deep}. It is well known \citep{neal1995bayesian,matthews2018gaussian} that in the limit of infinite width, NNs whose parameters are appropriately distributed converge to Gaussian processes. This connection helps to study properties of very wide NNs, and forms the basis of inferential algorithms directly targeting the infinite-dimensional setting \citep{lee2018deep,garriga-alonso2018deep,lee2019wide,arora2019exact}. Based on this recent literature, it is natural to ask whether it is possible to set an analogous useful connection between infinitely deep neural networks (IDNN) and stochastic processes. At a first glance, this correspondence might prove elusive. To see why, we now look at the literature on initialization schemes. Indeed there is a duality between initialization schemes and Bayesian NNs: an initialization scheme can be seen as a prior on the model parameters, thus inducing a prior on the NN. A NN at initialization may thus be viewed as a stochastic process indexed by depth, whose distribution is defined by a sequence of conditional distributions mapping from each layer to the next. Early works focused on stabilizing the variance of key quantities of interest across the layers of deep NNs \citep{glorot2010understanding,he2015delving}. More recent works \citep{poole2016exponential,schoenholz2017deep, hayou2019impact} consider the impact of initializations to the propagation of the input signal.

\begin{figure}
    \centering
    \hspace*{-0.5cm}\includegraphics[width=0.75\linewidth]{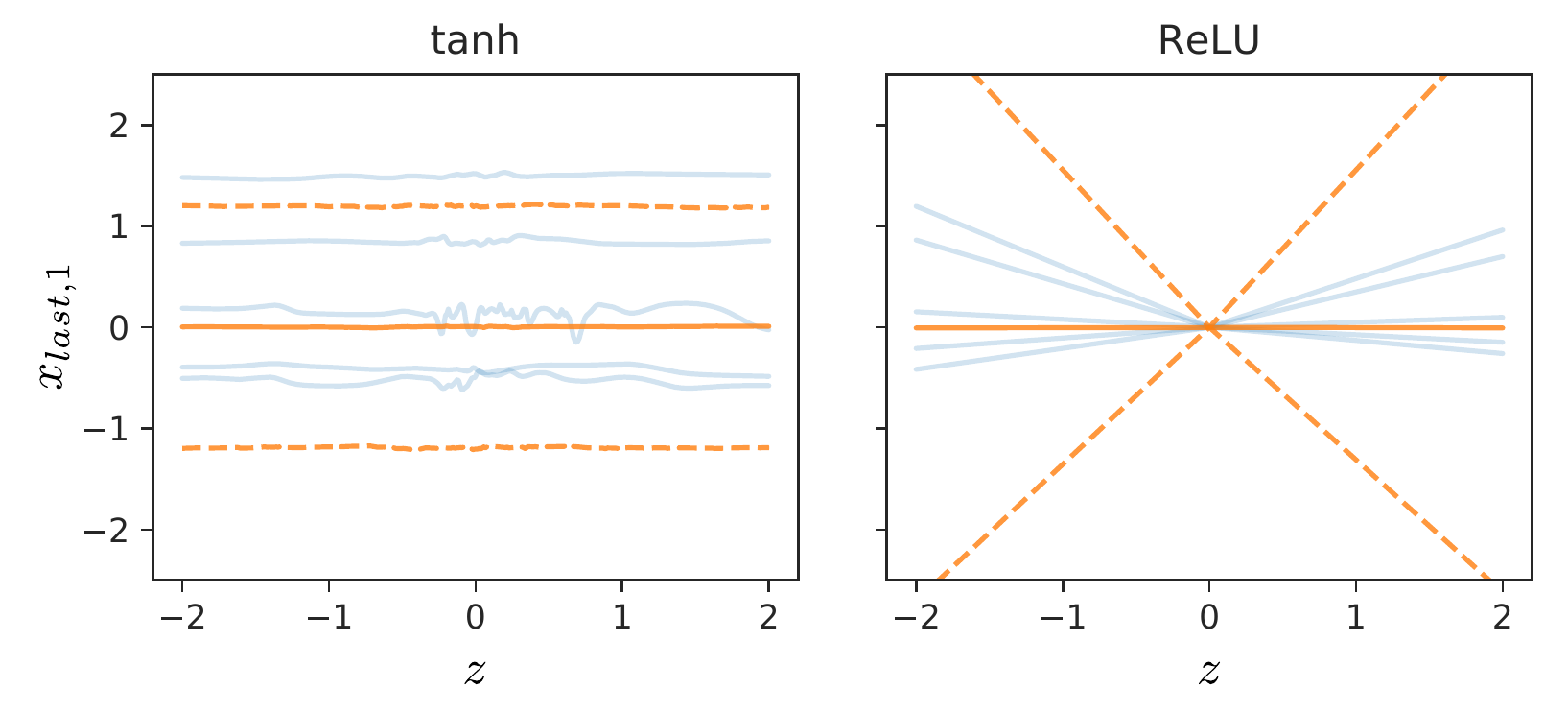}
    \caption{Function samples of a given pre-activation (number 1) of the last layer, $x_{last,1}$, of a fully connected feedforward NN with 500 layers of 500 units over a 1-dimensional input $z \in [-2,2]$;  $\tanh$ activation function and $\ReLU$ activation function, and parameters on the edge of chaos; 5 draws are displayed in blue in each figure; for each input the $5\%$, $50\%$ and $95\%$ quantiles are displayed in orange.}
    \label{fig:fspace_T_example}
\end{figure}

Even when initialized on the edge of chaos (EOC) for optimal signal propagation, feedforward NNs with fixed independent and identically distributed (i.i.d.) initialization exhibit some pathological properties as their total depth increases. In particular, the dependency  on the input eventually vanishes for most activation functions. In addition to that, the layers seen as random functions on the input space eventually concentrate on restrictive families including constant functions. As an illustrative example, we show in \cref{fig:fspace_T_example} function samples from the last layer of a feedforward deep NNs for two activation functions under EOC initialization. For a $\tanh$ activation, the input has no discernible impact on the output, as can be seen by the constant marginal distributions, and the sampled functions are almost constant. This behavior is representative of most smooth activation functions. For a $\ReLU$ activation, the input affects the variance of the output and the function samples are piece-wise linear. In both cases, the outputs of any two inputs end up perfectly correlated.
While this study applies to feedforward NNs, very deep residual networks (ResNet) suffer from similar issues \citep{yang2017mean}, with the additional issue that the variance of the Gaussian-distributed pre-activations may grow unbounded over layers.

While it is possible to obtain a \emph{well-defined} stochastic process corresponding to an IDNN, such a process is unexpressive: linear regression is a more flexible alternative. The difficulties discussed so far are determined by the fact that typical prior distributions on the model parameters introduce a constant level of randomness over each hidden layer. In this paper we consider prior distributions that depend on the number of layers, in such a way that they shrink as the number layers increases. This approach leads to our main result: as the number of layers increases, a class of ResNets converges, jointly over multiple inputs, to  diffusion processes on a finite time interval. The conditions required for attaining convergence provide us with a general guideline for selecting compatible NN architectures, activation functions and parameters distributions. The limiting diffusion processes satisfy suitable stochastic differential equations (SDE) that describe the evolution of IDNN layers over time (depth). The limiting diffusion is \emph{well-behaved} in the sense that: i) it retains dependency from the input; ii) it does not suffer from the perfect correlation constraint; iii) it does not collapse to a deterministic function nor does it diverge.

The paper is structured as follows. In \cref{sec:preliminaries_diffusion} we recall some preliminary results on diffusion limits of discrete-time stochastic process. \cref{sec:residual_diffusions} contains our main result: the convergence of a class of ResNets to solutions of SDEs. \cref{sec:numerical_experiments} contains numerical experiments and \cref{sec:conclusion} concludes. Proofs, additional experiments and plots, and additional discussions on related work are deferred to the Supplementary Material (SM).

\emph{Notation}: for a matrix $h$, $h\tran$ is its transpose, and if $h$ is square $\diag(h)$ is its diagonal vector and $\trace(h)$ is its trace; $\norm{x} = \sqrt{x\tran x}$ is the norm of the vector $x$; $\dprod{x}{y}=x \tran y$ is the inner product of vectors $x$ and $y$; $\norm{h} = \sqrt{\trace(h \tran h)}$ is the norm of a matrix $h$; $\vect(u)$ is the vectorization the tensor $u$; $\I$ is the identity matrix and $1$ is a vector of ones; for random variables $z$ and $w$, $\var[z]$, $\cov[z,w]$ and $\rho[z,w]$ are the variance, covariance and correlation; for random vectors $x \in \mathbb{R}^r$ and $y \in \mathbb{R}^c$, $\C[x,y]_{i,j} = \cov[x_i, x_j]$ is the $r \times c$ cross-covariance matrix $\C[x,y]$; $\V[x] = \C[x,x]$ is the $r \times r$ covariance matrix of $x$; the expectation $\E[u]$ of a random tensor $u$ is the tensor of the expectations of its elements; for two $D$-dimensional stochastic processes $x_t,y_t$,  $[x]_t$ is the quadratic variation (a $D$-dimensional vector) and $[x,y]_t$ is the quadratic covariation (a $D \times D$-dimensional matrix); $\1$ is the indicator function.

\section{Preliminaries} \label{sec:preliminaries_diffusion}

For $l=1,\dots,L$ let $\xdt_{l}$ be the $l$-th layer of a NN with with $L$ layers, and let $\xdt_0$ be the NN input. In this section we recall general results for diffusion approximations. The connection with NNs, i.e. defining what $\xdt_l$ exactly represents in a NN, is postponed to the next section. As we will be seeking a continuous time stochastic process limit we re-index $\xdt_0,\xdt_1,\dots,\xdt_L$ on a discrete time scale. Let $T > 0$ denote a terminal time, $\Dt = T/L$, for each $L$ we establish the correspondence between discrete indices $l \in \mathbb{Z}_{+}$ and discrete times $t \in \R_{+}$ by $l = 0,1,\dots,L \leftrightarrow t = 0,\Dt,2\Dt,\dots,T$. From now on we will consider without loss of generality a NN with input $\xdt_0$ and layers $\xdt_{\Dt},\dots,\xdt_T$, denoting a layer with $\xdt_t$.

Let $p(\xdt_T|\xdt_0)$ be the conditional distribution of the output given the input for a NN at initialization. Our strategy to enforce desirable properties on $p(\xdt_T|\xdt_0)$ consists in having a NN converge, as the number of layers $L$ go to infinity ($\Dt \downarrow 0$), to a continuous-time stochastic process on the time interval $[0,T]$. In this case, for $L$ large enough, the distribution $p(\xdt_T|\xdt_0)$ will be close to the distribution of the limiting process at terminal time $T$ given the same $\xdt_0$, and such limiting process should be chosen to make this transition density well behaved. In all NN architectures considered in this paper, each layer depends only on the previous one, hence $\xdt_t$ has the Markov property. These conditions identify a class of diffusion processes \citep{stroock2006multidimensional}, which are continuous-time Markov processes with continuous paths, as natural candidates for the limiting process. For simplicity we assume that the parameters of all layers follow the same distribution (extensions are discussed in \cref{sec:parameters_activations}), making $\xdt_t$ time-homogeneous.

Let $\xdt_t$ be a generic $D$-dimensional discrete-time Markov process and let $\Delta \xdt_t = \xdt_{t+\Dt} - \xdt_t$ define the forward increments. Hereafter we report a set of conditions that imply the convergence of $\xdt_t$ to the solution of a limiting SDE, and it is implicit that the distribution $p(\xdt_{t+\Dt}|\xdt_t)$ depends on $\Dt$ for the limits to exist as required.

\begin{assumption}[Convergence of instantaneous mean and covariance] \label{ass:inf_coeff}
There exist $\mu_x(x): \R^D \rightarrow \R^D$ and $\sigma_x^2(x): \R^D \rightarrow \R^{D \times D}$ such that:
\begin{align}
    &\lim_{\Dt \downarrow 0} \frac{\E[\Delta \xdt_t | \xdt_t]}{\Dt}   = \mu_x(\xdt_t)\label{eq:mu_x}\\
    &\lim_{\Dt \downarrow 0} \frac{\V[\Delta \xdt_t | \xdt_t]}{\Dt}   = \sigma_x^2(\xdt_t)\label{eq:sigma_x}\\
    &\lim_{\Dt \downarrow 0} \frac{\E[(\Delta \xdt_t)^{2 + \delta} | \xdt_t]}{\Dt} = 0\label{eq:continuity_x}
\end{align}
for some $\delta > 0$, where all convergences are uniform on compacts of $\mathbb{R}^D$ for each component, $\mu_x(x)$ and $\sigma_x^2(x)$ are continuous, and $\sigma_x^2(x)$ is positive semi-definite: $\sigma_x^2(x) = \sigma_x(x) \sigma_x(x)\tran$ for some $\sigma_x(x): \R^D \rightarrow \R^{D \times D}$.
\end{assumption}

Assumptions \eqref{eq:mu_x} and \eqref{eq:sigma_x} pinpoint the form of the limiting SDE, while assumption \cref{eq:continuity_x} is a technical condition that allows us to consider the limits \cref{eq:mu_x} and \cref{eq:sigma_x} instead of their truncated version \citep{nelson1990arch}.
The next theorem establishes that, under additional assumptions, in the limit $\xdt_t$ can be embedded in the solution of a SDE.

\begin{theorem} \label{thm:sde_convergence}
Under \cref{ass:inf_coeff}, extend $\xdt_t$ to a continuous-time process $\overline{\xdt}_t$ on $t \in [0,T]$ by continuous-on-right step-wise-constant interpolation of $\xdt_t$:
\begin{equation} \label{eq:sde_discrete_interpolated}
    \overline{\xdt}_t = \xdt_u\1_{u \leq t < u + \Dt}\qquad(u \in 0,\Dt,2\Dt,\dots,T)
\end{equation}
Consider the $D$-dimensional SDE on $[0,T]$ with initial value $x_0 = \xdt_0$, drift vector $\mu_x(x)$ given by \cref{eq:mu_x}, and diffusion matrix $\sigma_x(x)$ given by a square root of \cref{eq:sigma_x}:
\begin{equation} \label{eq:sde}
    dx_t = \mu_x(x_t) dt + \sigma_x(x_t) dB_t
\end{equation}
where $B_t$ is a $D$-dimensional Brownian motion (BM) with independent components and \cref{eq:sde} is short-hand notation for:
\begin{equation*}
    x_T = x_0 + \int_0^T \mu_x(x_t) dt + \int_0^T \sigma_x(x_t) dB_t
\end{equation*}
The first integral is a standard (Riemann) integral, and the second integral is an Ito integral. If  SDE \cref{eq:sde} admits a weak solution, and if this solution is unique in law and non-explosive, then the stochastic process defined by \cref{eq:sde_discrete_interpolated} converges in law to the solution of the SDE \cref{eq:sde}. This result still holds true for a random but independent and square integrable random variable $\xdt_0 \sim p(\xdt_0)$, provided that the driving BM is independent of $\xdt_0$. In both cases the convergence in law is on $\mathcal{D}([0,\infty),\mathbb{R}^D)$, the space of $\mathbb{R}^D$-valued processes on $[0,\infty)$ which are continuous from the right with finite left limits, endowed with the Skorohod metric \citep{billingsley1999convergence}.
\end{theorem}

We are dealing with three processes: the (discrete-time) NN $\xdt_t$, its continuous time interpolation $\overline{\xdt}_t$, and the limiting diffusion $x_t$ \citep{oksendal2003stochastic}. In Theorem \ref{thm:sde_convergence}, the continuous-time interpolation $\overline{\xdt}_t$ of $\xdt_t$ is introduced because we are seeking a continuous-time limiting process from a discrete-time one. The convergence established in Theorem \ref{thm:sde_convergence} is strong in the sense that it concerns the convergence of the distribution of the stochastic process $(\overline{\xdt}_t)_{t \in [0,T]}$ as a stochastic object on the whole time interval $[0,T]$ to the diffusion limit $(x_t)_{t \in [0,T]}$ as $L \uparrow \infty$. We consider weak solutions, as opposed to a strong ones, where it suffices that a BM $B_t$ can be found such that a solution can be obtained \citep{oksendal2003stochastic}. The focus on weak solutions and uniqueness in law of such solutions (also called weak uniqueness) is justified by our interest in the distributional properties of the limiting behavior of $\xdt_t$, and it enables us to consider weaker requirements for attaining convergence of $\xdt_t$. Consider the discretization of SDE \cref{eq:sde}
\begin{equation} \label{eq:euler_sde}
    x_{t+\Dt} = x_t + \mu_x(x_t) \Dt + \sigma_x(x_t) \zeta_t \sqrt{\Dt},
\end{equation}
where $\zeta_t$ is a $D$-dimensional random vector whose components are i.i.d. as standard Gaussian (mean $0$ and variance $1$). Under suitable conditions  \citep{kloeden1992numerical}, it can be proved that  the discretized SDE \cref{eq:euler_sde} converges to the SDE \cref{eq:sde}, and we recognize the Euler discretization of an ordinary differential equation (ODE) in the deterministic part \cref{eq:euler_sde}. In \cref{thm:sde_convergence} we postulate the existence and uniqueness in law of the weak solution of the limiting SDE, and its non-explosive behavior. The following conditions suffice for our goals.

\begin{assumption}[Existence of weak solution and uniqueness in law on compact sets] \label{ass:existence_uniqueness}
The functions $\mu_x(x)$ and $\sigma_x(x)$ are twice continuously differentiable.
\end{assumption}

\begin{assumption}[Non-explosive solution] \label{ass:non_explosivity}
There exist a finite $C > 0$ such that for each $x \in \R^{D}$: $\norm{\mu_x(x)} + \norm{\sigma_x(x)} \leq C(1 + \norm{x})$.
\end{assumption}

When \cref{ass:inf_coeff} and \cref{ass:existence_uniqueness} hold (as it will be the case in all the models considered), but \cref{ass:non_explosivity} does not hold, we still obtain convergence to the solution of the SDE \cref{eq:sde}. However, the stochastic process $x_t$ might diverge to infinity with positive probability on any time interval. We will return to this point more in detail.

\section{Residual network diffusions}\label{sec:residual_diffusions}

We focus on unmodified, albeit simplified, standard architectures. This is in line with the information propagation research \citep{poole2016exponential,schoenholz2017deep,hayou2019impact} but in contrast with \cite{chen2018neural}, where the recursion is modified with an additional $\Dt$ term to achieve convergence to a limiting ODE.

In this section, we study the implications of \cref{ass:inf_coeff}, \cref{ass:existence_uniqueness} and \cref{ass:non_explosivity} in NNs. First of all, $\xdt_t$ needs to be of constant dimensionality, as otherwise $\Delta \xdt_t$ is undefined. Consistently with the previous section we assume $\xdt_t \in \R^D$. For \cref{ass:inf_coeff} to hold we need $\Pr(\norm{\Delta \xdt_t}>\varepsilon|\xdt_t) \downarrow 0$ as $\Dt \downarrow 0$ for any $\varepsilon > 0$, i.e. we require the increments to vanish eventually. Intuitively this is due to the continuity of the paths of the limiting diffusion process. A fully connected feedforward NN is expressed by the relationship $\xdt_{t+\Dt} = f_t(\xdt_t) = \phi(A_t \xdt_t + a_t)$ for a nonlinear activation $\phi: \R \rightarrow \R$ applied element-wise. As standard convention we refer to $A_t \in \R^{D \times D}$ as weights and to $a_t \in \R^D$ as biases. Hence $\Delta \xdt_t = \phi(A_t \xdt_t + a_t) - \xdt_t$. Shrinking increments would imply that for all $x$, $\phi(A_t x + a_t)$ can be made arbitrarily concentrated around $x$ with a suitable choice of distributions for $(A_t,a_t)$. This cannot be achieved unless $\phi$ is linear or the distribution of $(A_t,a_t)$ depends on $x$. Indeed, fixing $x$ determines the values around which $(A_t,a_t)$ need to concentrate for the increments to vanish (if any), hence the increments will not vanish for a different $x' \neq x$, a fact that is most easily seen in the specific case where $(A_t,a_t)$ are scalars. The same reasoning rules out the ResNet originally introduced in the work of \cite{he2016deep}, where $\xdt_{t+\Dt} = f_t(\xdt_t + r_t(\xdt_t))$. This leaves us with the identity ResNet of \cite{he2016identity} where $\xdt_{t+\Dt} = \xdt_t + r_t(\xdt_t)$ for some choice of $r_t$, the residual blocks, which we require to eventually vanish.

\subsection{Shallow residual blocks}

Each residual block $r_t$ results from an interleaved application of affine transforms and non-linear activation functions. We consider the case of shallow residual blocks of the form:
\begin{equation}\label{eq:shallow_block}
    \xdt_{t+\Dt} = \xdt_t + \phi(A_t \psi(\xdt_t) +  a_t)
\end{equation}
for two activation functions $\phi: \R \rightarrow \R$, $\psi: \R \rightarrow \R$ which are applied element-wise. We point out that the non-standard use of 2 activation functions $\phi$, $\psi$ is to cover the case of shallow residual blocks in full generality.

\subsection{Parameter distribution and activation functions} \label{sec:parameters_activations}

For a shallow residual block $r_t$, the vanishing increments requirement is satisfied by having the distributions of $A_t$ and $a_t$ concentrate around 0 provided that $\phi(0) = 0$. It proves advantageous to consider weights and biases given by increments of diffusions corresponding to solvable SDEs.

\begin{assumption}[Parameters distribution and scaling] \label{ass:diffusion_parameters}
Let $W_t$ and $b_t$ be the diffusion processes respectively with values in $\R^{D \times D}$ and $\R^{D}$ solutions of:
\begin{align}
    &dW_t = \mu^W dt + d\widetilde{W}_t; \;\; d\vect(\widetilde{W}_t) = \sigma^W d\vect(B^W_t)\label{eq:param_diff_W}\\
    &db_t = \mu^b dt + \sigma^b dB^b_t\label{eq:param_diffusion_b}
\end{align}
where $B^W_t$ and $B^b_t$ are independent BMs with independent components respectively with values in $\R^{D \times D}$ and $\R^{D}$, $\mu^W \in \R^{D \times D}, \mu^b \in \R^D$, $\sigma^W \in \R^{D^2 \times D^2}, \sigma^b \in \R^{D \times D}$, and $\Sigma^W = \sigma^W {\sigma^W}\tran$, $\Sigma^b = \sigma^b {\sigma^b} \tran$ are positive semi-definite.
\end{assumption}

Then the discretizations of $W_t$ and $b_t$ admit the (exact) representations:
\begin{alignat*}{2}
    &\Delta W_t = \mu^W \Dt + \varepsilon^W_t \sqrt{\Dt}; \;\; &&\Delta b_t = \mu^b \Dt + \varepsilon^b_t \sqrt{\Dt}\\
    &\vect(\varepsilon^W_t) \overset{i.i.d.}{\sim} \mathcal{N}_{D^2}\big(0, \Sigma^W\big); \;\;  &&\varepsilon^b_t \overset{i.i.d.}{\sim} \mathcal{N}_{D}\big(0, \Sigma^b\big)
\end{alignat*}
for $t=\Dt,\dots,T$ where $\mathcal{N}$ stands for the multivariate Gaussian distribution.
We will consider residual blocks where $A_t = \Delta W_t$ and $a_t = \Delta b_t$:
\begin{equation}
    \xdt_{t+\Dt} = \xdt_t + \phi(\Delta W_t \psi(\xdt_t) + \Delta b_t)\label{eq:resnet_fc}
\end{equation}
Thus \cref{ass:diffusion_parameters} covers the case where the parameters are independently and identically distributed across layers according to an arbitrary multivariate Gaussian distribution, up to the required scaling which is necessary to obtain the desired diffusion limit. By considering deterministic but time-dependent $\mu^W_t,\mu^b_t,\Sigma^W_t,\Sigma^b_t$ the extension to layer-dependent distributions is immediate. More generally, we can consider $W_t$ and $b_t$ driven by arbitrary SDEs. Moreover, dependencies across the parameters of different layers can be accommodated by introducing additional SDE-driven processes, commonly driving the evolution of $W_t$ and $b_t$. We do not pursue further these directions in the present work. As for the activation functions, we will require:

\begin{assumption}[Activation functions regularity] \label{ass:activation}
The function $\phi: \R \rightarrow \R$ satisfies: $\phi(0) = 0$, $\phi$ is continuously differentiable three times on $\mathbb{R}$, its second and third derivatives have at most exponential tails growth, i.e. for some $k > 0$:
\begin{equation*}
    \lim_{|x| \uparrow \infty} \frac{|\phi''(x)|}{e^{k |x|}} + \lim_{|x| \uparrow \infty} \frac{|\phi'''(x)|}{e^{k |x|}} < \infty
\end{equation*}
The function $\psi: \R \rightarrow \R$ is locally bounded and continuously differentiable two times on $\R$.
\end{assumption}

\subsection{Diffusion limits}\label{sec:diffusion_limits}

The next theorem is the main result of the present paper, regarding the convergence of \cref{eq:resnet_fc}. Proofs are in SM A.

\begin{theorem} \label{thm:resnet_fc_sde}
Under \cref{ass:diffusion_parameters} and \cref{ass:activation}
the continuous-time interpolation $\overline{\xdt}_t$ of $\xdt_t$ converges in law to the solution on $[0,T]$ of
\begin{align} 
    &dx_t = \phi'(0) (\V[\varepsilon^W_t \psi(x_t) + \varepsilon^b_t|x_t])^{1/2} dB_t\label{eq:resnet_fc_sde}\\
    &\quad+ \phi'(0) (\mu^b + \mu^W \psi(x_t))dt\nonumber\\
    &\quad+ \frac{1}{2} \phi''(0) \diag(\V[\varepsilon^W_t \psi(x_t) + \varepsilon^b_t|x_t])dt\nonumber
\end{align}
with initial value $x_0 = \xdt_0$ where $B_t$ is a $D$-dimensional BM vector with independent components.
\end{theorem}

This result does not establish a direct connection between $x_t$ and the driving sources of stochasticity $W_t$ and $b_t$. As we are interested in the properties of deep ResNets in function space, i.e. over multiple inputs, a brute force approach would require us to establish diffusion limits as in \cref{thm:resnet_fc_sde} for an enlarged $\xdt_t=[\xdt_t^{(1)} \cdots \xdt_t^{(N)}] \in \R^{DN}$ corresponding to $N$ initial values $\xdt_0=[\xdt_0^{(1)} \cdots \xdt_0^{(N)}]$. Instead, we show that the limiting SDE is equivalent in law to the solution of another SDE which preserves the dependency on the driving sources of stochasticity. From here on $\xdt_t^{(i)},\xdt_t^{(j)}$ denote ResNets corresponding to two initial values $\xdt_0^{(i)},\xdt_0^{(j)}$, and $x_t^{(i)},x_t^{(j)}$ denotes diffusion limits corresponding to the same two initial values (i.e. $x_0^{(i)} = \xdt_0^{(i)},x_0^{(j)} = \xdt_0^{(j)})$. We will continue to use $\xdt_t$ for $\xdt_t^{(i)}$ and $x_t$ for $x_t^{(i)}$ when no confusion arises.

\begin{corollary} \label{thm:resnet_fc_sde_2}
Under the same assumptions of \cref{thm:resnet_fc_sde} the limiting process is also given by the solution on $[0,T]$ of:
\begin{align}
    &dx_t^{(i)} = \phi'(0)(dW_t \psi(x_t^{(i)}) + db_t)\label{eq:resnet_fc_sde_2}\\
    &\quad+ \frac{1}{2}\phi''(0)(d[W \psi(x^{(i)})]_t + d[b]_t)\nonumber
\end{align}
where $W_t$ and $b_t$ are defined in \cref{ass:diffusion_parameters} and over two initial values we have:
\begin{equation} \label{eq:resnet_fc_sde_2_cross}
    d[x^{(i)},x^{(j)}]_t = \phi'(0)^2(d[W\psi(x^{(i)}),W\psi(x^{(j)})]_t + d[b,b]_t)
\end{equation}
\end{corollary}

The results obtained so far are general in the sense that we allow for an arbitrary covariance structure between the elements of $\varepsilon^W_t$, i.e. an arbitrary (constant and deterministic) quadratic covariation for $W_t$. This makes it difficult to derive more explicit results, and is also an impractical approach as the parametrization requires $\mathcal{O}(D^4)$ elements. We thus consider more restrictive distribution assumptions with a more manageable $\mathcal{O}(D^2)$ parametrization cost.

\begin{assumption}[Matrix normal weights] \label{ass:diffusion_parameters_matrix_variate}
Let $b_t,\mu^b,\sigma_b,B^b_t,\mu^W,B^W_t$ be defined as in \cref{ass:diffusion_parameters}.
Let $W_t$ be the diffusion matrix with values in $\R^{D \times D}$ solution of:
\begin{equation*}
    dW_t = \mu^W dt + \sigma^{W_O} dB^W_t \sigma^{W_I}\\
\end{equation*}
where $\sigma^{W_O},\sigma^{W_I} \in \R^{D \times D}$ and $\Sigma^{W_O} = \sigma^{W_O} {\sigma^{W_O}} \tran$, $\Sigma^{W_I} = {\sigma^{W_I} }\tran \sigma^{W_I}$ are positive semi-definite.
\end{assumption}

Under \cref{ass:diffusion_parameters_matrix_variate} the discretization of $W_t$ satisfies:
\begin{equation*}
    \varepsilon^W_t \overset{i.i.d.}{\sim} \mathcal{MN}_{D,D}\big(0, \Sigma^{W_O}, \Sigma^{W_I} \big)
\end{equation*}
for $t=\Dt,\dots,T$ where $\mathcal{MN}$ stands for the matrix normal distribution. This is an immediate consequence of the fact that if $\zeta \sim \mathcal{MN}(0, \I, \I)$, then $A \zeta B \sim \mathcal{MN}(0, A A\tran, B\tran B)$. See \cite{gupta1999matrix}. The main property of $\mathcal{MN}$ distributions is that the covariance factorizes as $\cov(\varepsilon^W_{o,i},\varepsilon^W_{o',i'}) = \Sigma^{W_O}_{o,o'}\Sigma^{W_I}_{i,i'}$.

\begin{corollary} \label{thm:resnet_fc_matrix_variate}
Under the same assumptions of \cref{thm:resnet_fc_sde}, if $W_t$ is distributed according to \cref{ass:diffusion_parameters_matrix_variate}, \cref{eq:resnet_fc_sde_2} and \cref{eq:resnet_fc_sde_2_cross} are given by:
\begin{align}
    &dx_t^{(i)} = \phi'(0)\big((\mu^W \psi(x_t^{(i)}) + \mu^b) dt\label{eq:resnet_fc_matrix_variate}\\
    &\quad+ \sigma^{W_O} dB^W_t \sigma^{W_I} \psi(x_t^{(i)}) + \sigma^b dB^b_t\big)\nonumber\\
    &\quad+ \frac{1}{2}\phi''(0)\diag\big(\Sigma^b + \Sigma^{W_O} (\psi(x_t^{(i)})\tran \Sigma^{W_I} \psi(x_t^{(i)})) \big)dt\nonumber\\
    &d[x^{(i)},x^{(j)}]_t = \phi'(0)^2\big(\Sigma^b + \Sigma^{W_O} \psi(x^{(i)}_t)\tran \Sigma^{W_I} \psi(x^{(j)}_t)\big)dt\nonumber
\end{align}
\end{corollary}

Finally, we consider the simplest "fully i.i.d." centered distribution assumptions for $W_t$, $b_t$. i.i.d. initializations are most commonly used in the training of NNs. We also introduce a scaling of the weights by $D^{-1/2}$ (which is the same scaling used to obtain Gaussian process limits in infinitely wide NNs). We will see in \cref{sec:function_space_dist} that this scaling has a stabilizing effect on the dynamics of $x_t$.

\begin{assumption}[Fully i.i.d. parameters] \label{ass:diffusion_parameters_iid}
Let $W_t$ and $b_t$ be the diffusion processes respectively with values in $\R^{D \times D}$ and $\R^{D}$ solutions of:
\begin{equation*}
    dW_t = \frac{\sigma_w}{\sqrt{D}} dB^W_t; \;\; db_t = \sigma_b dB^b_t
\end{equation*}
for $B^W_t,B^b_t$ independent BMs respectively with values in $\R^{D \times D},\R^D$ and scalars $\sigma_w>0,\sigma_b>0$.
\end{assumption}

Under \cref{ass:diffusion_parameters_iid} the discretizations of $W_t,b_t$ satisfy:
\begin{alignat}{2}
    &\Delta W_t = \varepsilon^W_t \frac{\sigma_w}{\sqrt{D}} \sqrt{\Dt}; \;\; &&\Delta b_t = \varepsilon^b_t \sigma_b \sqrt{\Dt}\label{eq:fc_iid_discr_std}\\
    &\varepsilon^W_t \overset{i.i.d.}{\sim} \mathcal{MN}_{D,D}\big(0, \I_D, \I_D \big); \;\; &&\varepsilon^b_t \overset{i.i.d.}{\sim} \mathcal{N}_{D}\big(0, \I_D \big)\label{eq:fc_iid_discr_repar}
\end{alignat}

\begin{corollary} \label{thm:resnet_fc_iid}
Under the same assumptions of \cref{thm:resnet_fc_sde}, if $W_t$ and $b_t$ are distributed according to \cref{ass:diffusion_parameters_iid}, \cref{eq:resnet_fc_sde_2} and \cref{eq:resnet_fc_sde_2_cross} are given by:
\begin{align}
    &dx_t^{(i)} = \phi'(0)\big(\frac{\sigma_w}{\sqrt{D}} \norm{\psi(x_t^{(i)})} dB^W_t  + \sigma_b dB^b_t\big)\label{eq:resnet_fc_iid}\\ 
    &\quad+ \frac{1}{2}\phi''(0)\big(\sigma_b^2 + \frac{\sigma_w^2}{D} \norm{\psi(x_t^{(i)})}^2) \big) \I_D dt\nonumber\\
    &d[x^{(i)},x^{(j)}]_t = \phi'(0)^2\big(\sigma_b^2 + \frac{\sigma_w^2}{D} \dprod{\psi(x^{(i)}_t)}{\psi(x^{(j)}_t)}\big) \I_D dt\nonumber
\end{align}
\end{corollary}

\subsection{Qualitative properties}

\emph{Non-vanishing input dependency:} a consequence of \cref{thm:resnet_fc_sde} is that the distribution of the ResNet output given the input $p(\xdt_T|\xdt_0)$ converges to the transition density $p(x_T|x_0)$ of the solution of \cref{eq:resnet_fc_sde_2}. As $T$ is finite, the dependency on the input does not vanish in the limit of infinite total depth $L$ and can be controlled via the parameter distributions and $T$.

\emph{Flexible output distributions:} from \cref{eq:resnet_fc_sde_2}-\cref{eq:resnet_fc_sde_2_cross} we see that the joint evolution of $x_t^{(i)},x_t^{(j)}$ corresponding to $x_0^{(i)},x_0^{(j)}$ is not perfectly correlated (unless there are no weight parameters, a not very relevant case). This remains true also in the parameterizations of \cref{ass:diffusion_parameters_matrix_variate} and \cref{ass:diffusion_parameters_iid}. Thus in the limit of infinite total depth $L$ the distribution in function space does not suffer from the perfect correlation problem. The joint distribution $p(x_T^{(i)},x_T^{(j)}|x_0^{(i)},x_0^{(j)})$ is not Gaussian.

\emph{Role of integration time:} a standard time-change result for SDEs \citep{revuz1999continuous} implies that time-scaling a SDE is equivalent to multiplying the drift and diffusion coefficients respectively by the scaling constant and by the square root of the scaling constant, as can be intuitively seen from \cref{eq:euler_sde}. From \cref{eq:resnet_fc_sde} we see that it is possible to compensate changes in the integration time $T$ with changes in the "hyper-parameters" $\mu^b,\mu^W,\Sigma^b,\Sigma^W$ in \cref{ass:diffusion_parameters} to leave the dynamics of \cref{eq:resnet_fc_sde} invariant. This remains true also in the parameterizations of \cref{ass:diffusion_parameters_matrix_variate} and \cref{ass:diffusion_parameters_iid}.
Hence we can restrict $T=1$ without loss of generality.

\emph{Matrix normal weights:} in this case $\V[\varepsilon^W_t \psi(x_t) + \varepsilon^b_t|x_t]$ is given by $\Sigma^b + \Sigma^{W_O} (\psi(x_t)\tran \Sigma^{W_I} \psi(x_t))$. The dependency on the state $x_t$ in \cref{eq:resnet_fc_sde} goes through a linear transformation and a weighted inner product. This sheds some light on the impact of introducing dependencies among row and columns of the weight parameters $A_t = \Delta W_t$. Specifically, $\Sigma^{W_I}$ define the structure of the inner weighted product, while $\Sigma^{W_O}$ defines how such transforms affect each dimension $d \in D$.

\emph{Fully i.i.d. parameters:} in this case $\V[\varepsilon^W_t \psi(x_t) + \varepsilon^b_t|x_t]$ is given by $\sigma_b^2 + \frac{\sigma_w^2}{D} \norm{\psi(x_t)}^2$. The dependency on the state $x_t$ in \cref{eq:resnet_fc_sde} goes only through the norm of $x_t$ which is permutation invariant in $d \in D$. Thus the law of the processes $x_{t,d}$ is exchangeable across $d \in D$ if the distribution of $x_{0,d}$ is so.

\emph{Explosive solutions:} without further assumptions the solutions to the limiting SDEs can be explosive. From \cref{eq:resnet_fc_sde} we see that the potentially troublesome term is the variance matrix in the drift (\cref{eq:resnet_fc_matrix_variate} makes the issue easier to see in a more restricted setting). \cref{ass:non_explosivity} is satisfied under all considered parameter distribution assumptions if either: i) $\psi$ exhibits at most square-root growth, in particular $\psi$ is bounded; or ii) $\psi$ exhibits at most linear growth, in particular $\psi$ is the identity function, and $\phi''(0)=0$, in particular $\phi = \tanh$.

\emph{Non-smooth activations:} 
the diffusion limits are based on a sufficiently smooth activation $\phi$ per \cref{ass:activation}. We consider here the following case which includes the ReLU activation. If $\phi(a)$ is positively homogeneous, i.e. $\phi(\alpha a) = \alpha \phi(a)$ for $\alpha > 0$, $h$ is random variable, and $\gamma > 0$ then: $\E[\phi(h \Dt^\gamma)/\Dt] = \E\left[\phi(h) \right] \Dt^{\gamma - 1}$ and $\E[\phi(h \Dt^\gamma)^2/\Dt] = \E\left[\phi(h)^2\right] \Dt^{2\gamma - 1}$. Comparing these with \cref{eq:mu_x} and \cref{eq:sigma_x}, we see that unless $\E[\phi(h)] = 0$, choosing $\gamma = 1/2$ would result in the drift term blowing up. Choosing $\gamma = 1$ recovers a deterministic limit as in \cite{chen2018neural}.

\subsection{Input and output layers}\label{sec:input_output}

So far we have considered $\xdt_0 \in \R^D$ to be the input of the ResNet. A NN acts as a function approximator to be fitted to some dataset $\{(z^{(i)},y^{(i)})\}_{i=1}^N$ where $z^{(i)} \in \R^Z$ represents an input and $y^{(i)} \in \R^Y$ represents the corresponding output. In general, there can be a mismatch between $D,Z$ and $Y$, making it is necessary to introduce adaptation layers $z^{(i)} \mapsto \xdt_0^{(i)}$ and $\xdt_T^{(i)} \mapsto \widehat{y}^{(i)}$ where $\widehat{y}^{(i)}$ is the NN prediction for $z^{(i)}$. As for $\xdt_t$, we will denote a single data-point $(z^{(i)},y^{(i)})$ with $(z,y)$ when no confusion arises.

\section{Experiments}\label{sec:numerical_experiments}

\subsection{Sanity check} \label{sec:sanity_check}
\begin{figure}
    \centering
    \includegraphics[width=\linewidth]{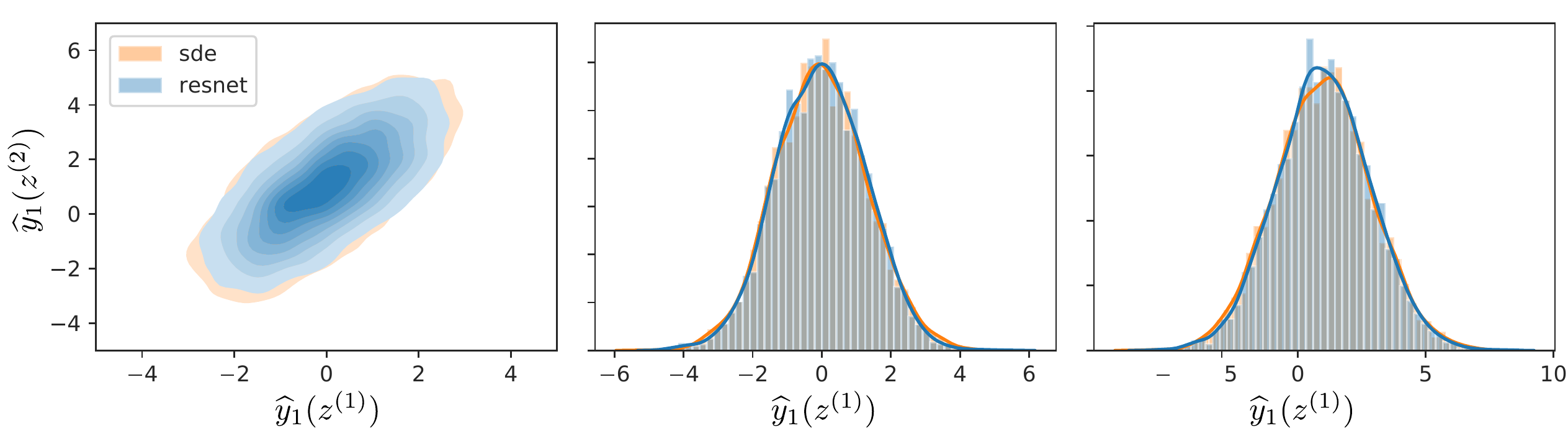}
    \caption{For model $\mathcal{SC}_{\tanh}$: 2D KDE plot for $(\widehat{y}_1(z^{(1)}),\widehat{y}_1(z^{(2)}))$ (left), 1D KDE and histogram plots for $\widehat{y}_1(z^{(1)})$ (center), $\widehat{y}_1(z^{(2)})$ (right) when $\widehat{y}_1$ is sampled from a ResNet and from the Euler discretization of its limiting SDE (sde); $\widehat{y}$ denotes a generic model output, hence $\widehat{y}_1$ is its first dimension.}
    \label{fig:sanity_check}
\end{figure}

First of all we investigate numerically the correctness of the results obtained in \cref{sec:diffusion_limits}. We consider the setting of \cref{ass:diffusion_parameters_iid} with $\phi = \tanh$, $\sigma_w^2=\sigma_b^2=1$, $T=1$, $L=D=500$ and 1-dimensional inputs. In all the experiments $\psi$ is set to the identity function. As noted in \cref{sec:input_output} we need to introduce an input layer mapping $z \in \R \mapsto \xdt_0 \in \R^D$. For this toy example we simply copy the input across all dimensions: $\xdt_{0,\bullet} = z$, i.e. $\xdt_{0,d} = z$ for each $d \in D$. We refer to this model as $\mathcal{SC}_{\tanh}$. We consider two inputs $z^{(1)} = 0$, $z^{(2)} = 1$, hence $\xdt^{(1)}_{0,\bullet} = z^{(1)},\xdt^{(2)}_{0,\bullet} = z^{(2)}$, and simulate $10.000$ draws of the first dimension ($d = 1$) of i) $\xdt^{(1)}_T$, $\xdt^{(2)}_T$ via the ResNet recursion \cref{eq:shallow_block}; ii) $x^{(1)}_T$, $x^{(2)}_T$ via the discretization \cref{eq:euler_sde} of the limiting SDE \cref{eq:resnet_fc_iid}. Our analysis imply that i) and ii) are equivalent in the limit $L \uparrow \infty$. We report the results in \cref{fig:sanity_check} where good agreement is indeed observed. We replicate this experiment in SM B for $\mathcal{SC}_{\swish}$, where the $\tanh$ activation in $\mathcal{SC}_{\tanh}$ is replaced by the $\swish$ activation ($\swish(x) = x \sigmoid(x)$) which has been shown empirically \citep{ramachandran2017searching} and theoretically \citep{hayou2019impact} to be competitive. In this case $\phi'(0) = \phi''(0) = 1/2$ and \cref{ass:non_explosivity} is not satisfied.

\subsection{Function space distributions}\label{sec:function_space_dist}

We show empirically that the dependency on the input is retained and the output distribution does not exhibit perfect correlation for very deep ResNet constructed as in the present paper. We consider the same model $\mathcal{SC}_{\tanh}$ of \cref{sec:sanity_check}. First of all, from the center and right plots of \cref{fig:sanity_check} we see that $\xdt^{(1)}_{T,1}$ and $\xdt^{(2)}_{T,1}$ are differently distributed, meaning the input dependency is retained, and from the left plot we see that they are not perfectly correlated, otherwise the 2D KDE would collapse to a straight line.

\begin{figure}
    \centering
    \includegraphics[width=0.7\linewidth]{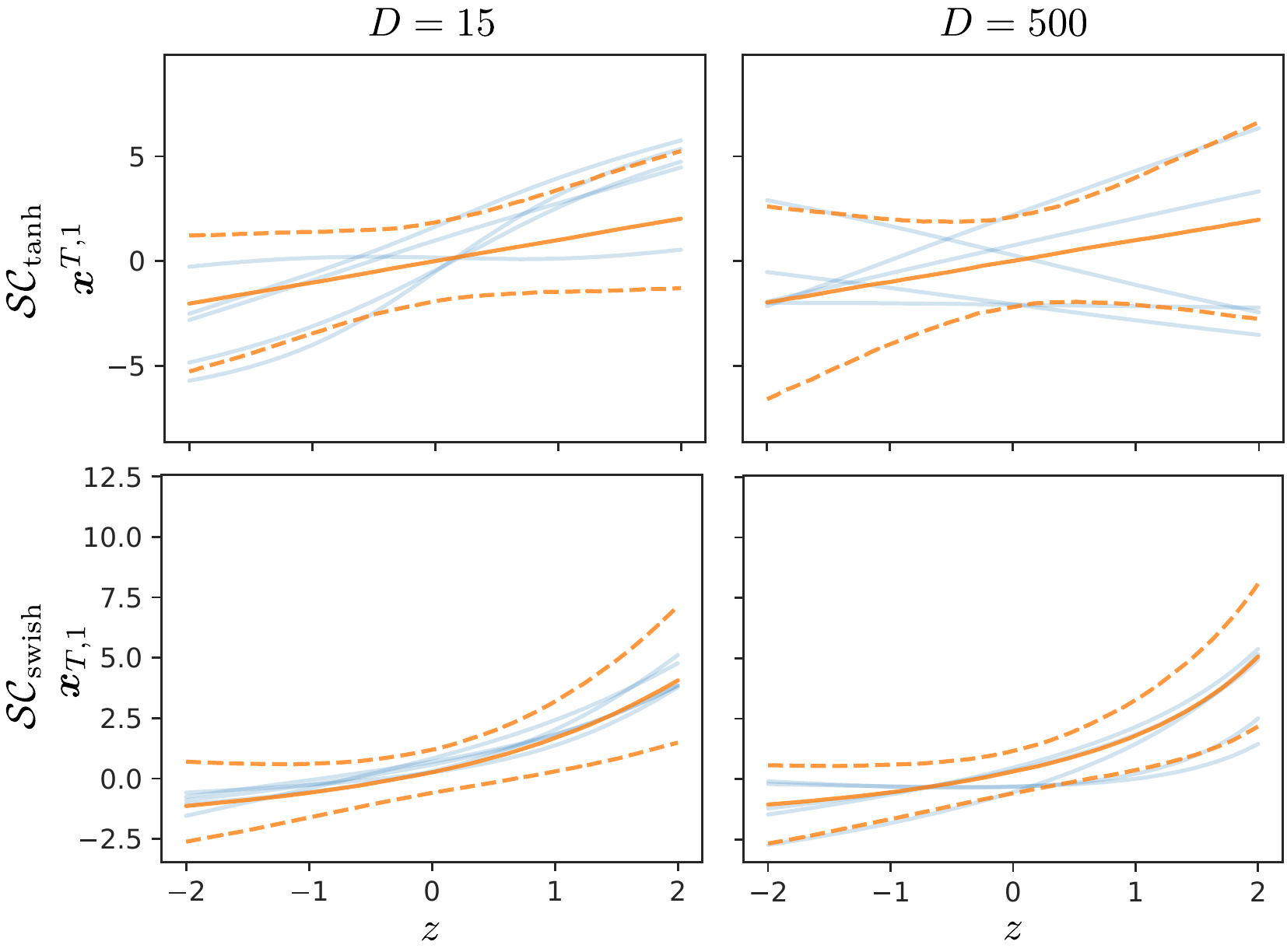}
    \caption{Function samples of $\xdt_{T.1}$ for $\mathcal{SC}_{\tanh}$ (top) and ${SC}_{\swish}$ (bottom), see \cref{fig:fspace_T_example} for the description of the plotted quantities.}
    \label{fig:f_diffusion}
\end{figure}

In \cref{fig:f_diffusion} (top) we visualize samples of $\xdt_{T,1}$ from $\mathcal{SC}_{\tanh}$ in function space for different combinations of $L$ (more plots in SM B). More specifically, we approximate function draws by considering 400 inputs $z^{(i)}$ equally spaced on $[-2,2]$. Using the ResNet recursion \cref{eq:shallow_block} we obtain 400 output values $\xdt_{T,1}^{(i)}$. We repeat this procedure to obtain 10.000 function draws.
In \cref{fig:f_diffusion} (bottom) we repeat this experiment for $\mathcal{SC}_{\swish}$. In this specific case we did not observe divergent trajectories for the $10.000$ function draws. In \cref{fig:f_diffusion} we observe similar distribution properties across different orders of magnitude for $D$, which suggests the existence of a stochastic limit in the doubly infinite setting where $L,D \uparrow \infty$.

\begin{figure}
    \centering
    \includegraphics[width=0.7\linewidth]{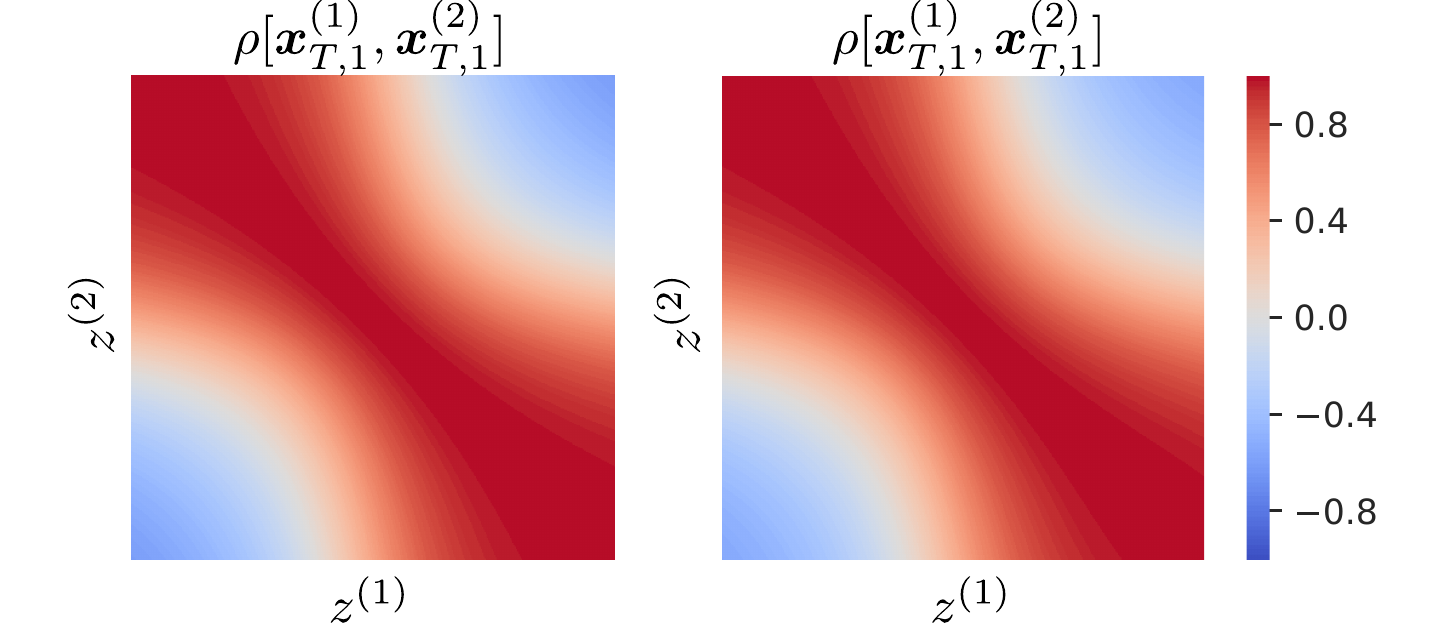}
    \includegraphics[width=0.7\linewidth]{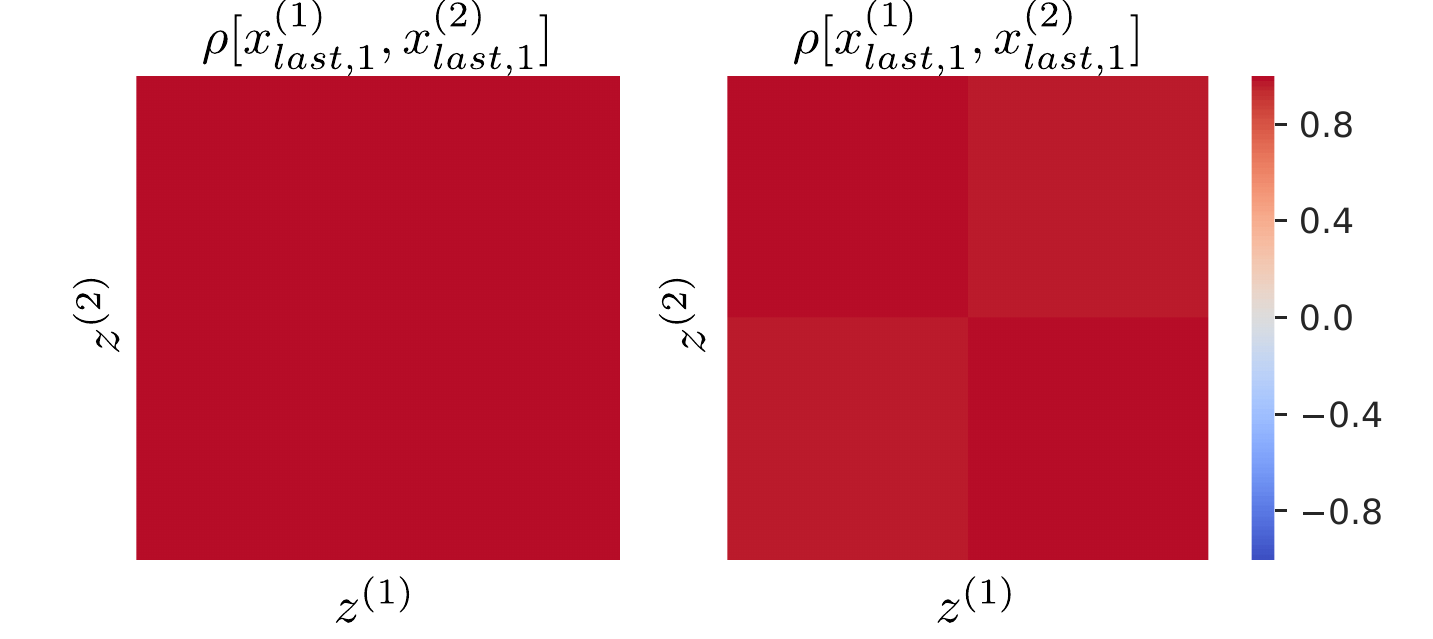}
    \caption{Output correlation heatmap for $\mathcal{SC}_{\tanh}$ (top-left), $\mathcal{SC}_{\swish}$ (top-right),  $\mathcal{EO}_{\tanh}$ (bottom-left), $\mathcal{EO}_{\ReLU}$ (bottom-right).}
    \label{fig:corr_sde}
\end{figure}

In \cref{fig:corr_sde} (top) we plot the correlations $\rho[\xdt^{(1)}_{T,1},\xdt^{(2)}_{T,1}]$ for inputs $(z^{(1)},z^{(2)})$ in the range $[-2,2] \times [-2,2]$ for the $\tanh$ and $\swish$ activations: for different inputs the output correlations are far from 1. Let us refer to the model of \cref{fig:fspace_T_example} with $\tanh$ activation as $\mathcal{EO}_{\tanh}$, and to the model of \cref{fig:fspace_T_example} with $\ReLU$ activation as $\mathcal{EO}_{\ReLU}$. For comparison, we show in \cref{fig:corr_sde} (bottom) the correlations $\rho[x^{(1)}_{last,1},x^{(2)}_{last,1}]$ for pre-activation 1 for $\mathcal{EO}_{\tanh}$ and $\mathcal{EO}_{\ReLU}$: all correlations are close to 1.

\subsection{SGD training}\label{sec:sgd}

\begin{figure}
    \centering
    \includegraphics[width=0.85\linewidth]{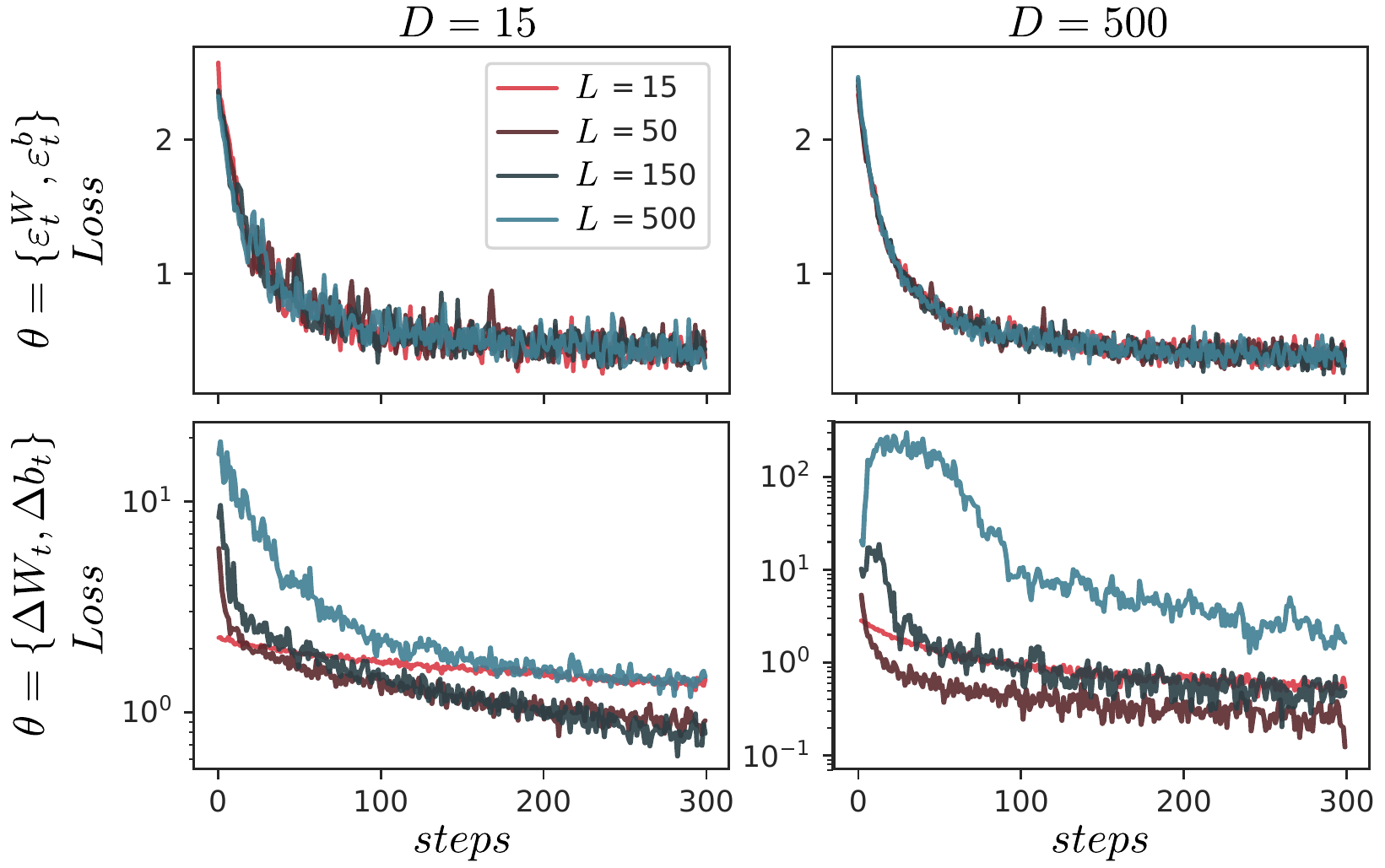}
    \caption{Averaged (over each batch) loss on MNIST training dataset for the model of \cref{sec:sgd}, different $L,D$, for reparametrized gradients (top, shared linear-scale on $y$-axes) and for standard gradients (bottom, different log-scales on $y$-axes).}
    \label{fig:sgd}
\end{figure}

In this experiment we consider the MNIST dataset \citep{lecun1998mnist}. Each observation $(z,y)$ is composed of an image $z \in \R^{784}$ (we flatten to a vector) and a class $y \in \R^{10}$ (we use 1-hot encoding). We consider the setting of \cref{ass:diffusion_parameters_iid} with $\phi = \tanh$, $\sigma_w^2=\sigma_b^2=1$, $T=1$ and random input and output layers given by $\xdt_0 = W_I z, \widehat{y} = W_O \xdt_T$ where $W_I \in \R^{D \times 784}, W_O \in \R^{10 \times D}$ and $W_{I,d,i},W_{O,c,o} \overset{i.i.d.}{\sim} \mathcal{N}(0,1)$. We use the cross-entropy loss function and fit the model to the training dataset via SGD. \cref{fig:sgd} (top) shows the evolution of the training losses over 1 epoch (mini-batches of 200 samples) when the gradients are taken with respect to $\{\varepsilon^W_t,\varepsilon^b_t\}_{t=0}^{T-\Dt}$ (\cref{eq:fc_iid_discr_repar}, reparametrized gradients) for a common learning rate.
This choice results in stable loss decrease over all considered values for $L$ and $D$. Moreover all average accuracies computed on the test dataset after 1 training epoch are in the range $[87.1\%, 90.6\%]$. In contrast, we were unable to obtain a test accuracy uniformly above $72.4\%$ with a common (tuned via grid-search) learning rate when the gradients are computed with respect to $\{\Delta W_t,\Delta b_t\}_{t=0}^{T-\Dt}$  (\cref{eq:fc_iid_discr_std}, standard gradients). \cref{fig:sgd} (bottom) illustrates the issue: a common learning rate leads to either slow or divergent trajectories. Similar results (not shown) are obtained for commonly used initializations (\cite{glorot2010understanding,he2015delving}). Our experiment suggests the existence of results akin to \cite{jacot2018neural, hayou2019training} as both $L,D \uparrow \infty$.

\cite{zhang2018residual} considers initializations for ResNets which are not encompassed yet by our analysis. Conversely, the residual blocks in \cite{zhang2018residual} cannot be shallow. An analysis of gradient properties motivates initializing the residual block parameters so that the their variance shrinks as the ResNet gets deeper. However, the residual blocks are multiplied by parameters initialized at 0, hence our desiderata iii) (\cref{sec:introduction}) is not satisfied. Moreover the gradients are not reparametrized as in the above experiment.

\section{Discussion}\label{sec:conclusion}

We have established the convergence of identity ResNets \cite{he2016identity} to solutions of SDEs as the number of layers goes to infinity. Our results rely on smooth activation functions and on model parameter distributions which shrink as total depth increases. Further conditions on the activation functions are obtained by restricting the limiting SDEs to be non explosive. As the infinitesimal evolution of SDEs is characterized by their instantaneous mean and covariance, it seemed natural to assume that model's parameters have Gaussian distributions. However, our results can be strengthened to hold for finite-variance parameter distributions.

Building on the connection between IDNN and diffusion processes we showed that, as the number of layers goes to infinity: the last layer does not collapse to a deterministic limit, nor does it diverge to infinity; the dependency of the last layer on the input does not vanish; the last layer, as stochastic function on input space, remains flexible without collapsing to restrictive families of distributions. We then investigated additional properties of the limiting diffusions. In contrast to the information propagation approach our analysis covers finitely-wide NNs and correlated parameters at the layer level.

While the limiting diffusions do not suffer from catastrophic limitations, to obtain competitive performance more attention needs to be paid to architectural choices, to parameters' distribution selection, and to input and output layers. Moreover, results on forward propagation do not trivially translate to corresponding results on gradient back-propagation. With this in mind, hereafter we list some promising future research directions. Firstly, we can consider more realistic residual blocks consisting of multiple convolutional layers as in \cite{zhang2018residual}. Extending the present work to convolutional NN does not require new theoretical developments as a convolutional transform (jointly over all positions) can be expressed via matrix multiplication. Deep residual blocks could be approached via fractional Brownian motions \citep{biagini2008stochastic} or via re-scaled Brownian motions. Secondly, the same techniques used to derive the evolution of IDNNs can be used to obtain the evolution of the input-output Jacobian. This would pave the way to an extensions of the neural tangent kernel \citep{jacot2018neural,lee2019wide,arora2019exact,hayou2019training} to IDNNs. Thirdly, stable behavior has been observed with an appropriate scaling of the weight parameters as the wideness $D$ increases. In particular, it would be instructive to characterize the distribution of NNs which are both infinitely deep and wide. This result could form the basis of Bayesian inference \citep{lee2018deep,garriga-alonso2018deep} for doubly infinite NNs and of data-dependent initializations.

\section{Acknowledgements}\label{sec:acknowledgements}

We wish to thank the three anonymous reviewers and the meta reviewer for their valuable feedback. The authors acknowledge Thierry Sousbie and Tiago Ramalho for the many suggestions that greatly improved the presentation of the current work. Stefano Favaro received funding from the European Research Council (ERC) under the European Union's Horizon 2020 research and innovation programme under grant agreement No 817257. Stefano Favaro gratefully acknowledge the financial support from the Italian Ministry of Education, University and Research (MIUR), ``Dipartimenti di Eccellenza" grant 2018-2022.

\clearpage
\bibliographystyle{apalike}
\bibliography{refs-new}

\end{document}


\onecolumn
\appendix

\section{Proofs} \label{app:proofs}

This section contains all the proofs of the theorems stated in the main text and the lemmas required to prove them.

\begin{proof}[Proof of Theorem 2.1]
This is \cite[Theorem 2.2]{nelson1990arch}: Assumption 2.1 and the postulated weakly unique and non-explosive weak solution satisfy all the conditions required for the application of \cite[Theorem 2.2]{nelson1990arch}.
Note that we use a stronger non-explosivity condition \cite{oksendal2003stochastic}.
Alternatively, for this standard result the reader can refer to the monograph \cite{stroock2006multidimensional} on which \cite{nelson1990arch} is based; yet another reference is \cite{ethier2009markov}.
\end{proof}

\begin{lemma} \label{lem:bounds}
If $\phi$ satisfies Assumption 3.2, $\epsilon \sim \mathcal{N}(0, \sigma^2)$ with $\sigma^2 \leq \sigma_*^2$, $\alpha > 0$, then we can find $M_2(\alpha, \sigma_*^2) < \infty$ and $M_3(\alpha, \sigma_*^2) < \infty$ such that:
\begin{align*}
    \E \left[ |\phi''(\epsilon)|^\alpha \right]  &\leq M_2(\alpha, \sigma_*^2) \\ 
    \E \left[ |\phi'''(\epsilon)|^\alpha \right] &\leq M_3(\alpha, \sigma_*^2)
\end{align*}
\end{lemma}
\begin{proof}
We prove the result only for $\phi''(\epsilon)$, the case for $\phi'''(\epsilon)$ being identical.
Let $L$ large enough such that $|\phi''(x)| \leq K_1 e^{K_2 |x|}$ for $|x| \geq L$ then:
\begin{align*}
    \E \left[ |\phi''(\epsilon)|^\alpha \right] &= \E \left[ |\phi''(\epsilon)|^\alpha \1_{|\epsilon| \leq L} \right] + \E \left[ |\phi''(\epsilon)|^\alpha \1_{|\epsilon| > L} \right] \\
                                                   &\leq \sup_{|x| \leq L}|\phi''(x)|^\alpha + K_1^\alpha \E[e^{K_2 \alpha |\epsilon|}]
\end{align*}
The first term is finite, that the second one can be bounded by a finite and increasing function in $\sigma^2$ follows from the symmetry in law of $\epsilon$ and the form of its movement generating function.
\end{proof}

\begin{proof}[Proof of Theorem 3.1]
We suppress the dependency on $t$ of vector and matrices and the conditioning in expectations and covariances in this proof to ease the notation. We also drop the boldness of $\xdt_t$ as no confusion arises in this setting. We instead reserve subscripts for indexing: for example $x_d$ denotes the $d$-th element of a vector $x$.

Let $h = (\mu^W \sqrt{\Dt} + \varepsilon^W) \psi(x) + (\mu^b \sqrt{\Dt} + \varepsilon^b)$ so that $h \sqrt{\Dt} = \Delta W \psi(x) + \Delta b$. By second order Taylor expansion of $\phi$ around 0 we have for $d = 1, \dots, D$
\begin{equation*}
     \frac{\Delta x_d}{\Dt} = \frac{\phi(h_d \sqrt{\Dt})}{\Dt} = \phi'(0) h_d \Dt ^ {-1/2} + \frac{1}{2} \phi''(0) h_d ^ 2 + \frac{1}{6} \phi'''(\vartheta_d) h_d ^ 3 \Dt ^ {1/2}
\end{equation*}
with $\vartheta_d \in (-h_d \sqrt{\Dt}, h_d \sqrt{\Dt})$. To prove (1) we want to show that $\forall R > 0$
\begin{equation*}
    \lim_{\Dt \downarrow 0} \sup_{\norm{x} < R} \left| \mu_x(x)_d - \E\left[\phi'(0) h_d \Dt ^ {-1/2} + \frac{1}{2} \phi''(0) h_d ^ 2 + \frac{1}{6} \phi'''(\vartheta) h_d ^ 3 \Dt ^ {1/2} \right] \right| = 0.
\end{equation*}
Now, $h_d = (\mu^W_d \sqrt{\Dt} + \varepsilon^W_d) \psi(x) + \mu^b_d \sqrt{\Dt} + \varepsilon^b_d$ and the distribution assumptions on $\varepsilon^W$ and $\varepsilon^b$ lead to
\begin{align*}
    \E\left[\phi'(0)h_d\Dt^{-1/2} + \frac{1}{2} \phi''(0) h_d ^ 2 \right] &= \phi'(0)(\mu^b_d + \mu^W_d \psi(x))\\
    &+ \frac{1}{2} \phi''(0) \V[\varepsilon^W \psi(x) + \varepsilon^b]_{d,d}\\
    &+ \frac{1}{2} \phi''(0) \left(\mu^b_d + \mu^W_d \psi(x)\right)^2\Dt \\
    &= \mu_x(x)_d + \frac{1}{2} \phi''(0) \left(\mu^b_d + \mu^W_d \psi(x)\right)^2\Dt.
\end{align*}
It remains to show that
\begin{equation*}
    \lim_{\Dt \downarrow 0} \sup_{\norm{x} < R} \left| \left(\mu^b_d + \mu^W_d \psi(x)\right)^2 \right| \Dt = 0,
\end{equation*}
which holds as $\psi$ is locally bounded, and that
\begin{equation*}
    \lim_{\Dt \downarrow 0} \sup_{\norm{x} < R} \left| \E\left[ \phi'''(\vartheta_d) h_d ^ 3 \right] \right| \Dt ^ {1/2} = 0,
\end{equation*}
for which it suffices to show that $\sup_{\norm{x} < R} \left| \E\left[ \phi'''(\vartheta_d) h_d ^ 3 \right] \right|$ can be bounded by $M(R) < \infty$ uniformly in $\Dt$. By Cauchy–Schwarz
\begin{equation*}
    \sup_{\norm{x} < R} \left| \E\left[ \phi'''(\vartheta_d) h_d ^ 3 \right] \right| \leq \sup_{\norm{x} < R} \E\left[ \phi'''(\vartheta_d) ^2 \right]^{1/2} \sup_{\norm{x} < R} \E\left[h_d ^ 6 \right]^{1/2}.
\end{equation*}
Again, as $\psi$ is locally bounded the constraint $\sup_{\norm{x} < R}$ corresponds to a constraint on the variance of $h_d$ hence the second $\sup$ is finite. By \cref{lem:bounds} the first $\sup$ is finite too and not increasing in $\Dt$ as $|\vartheta_d| \leq \sqrt{\Dt}|h_d|$ which allows us to produce the desired bound $M(R)$.

Regarding (3), by first order Taylor expansion of $\phi$ around 0 we need to show that for $d = 1, \dots, D$ and $R>0$
\begin{equation*}
    \lim_{\Dt \downarrow 0} \sup_{\norm{x} < R} \left| \E\left[ \frac{\left(\phi'(0) h_d \Dt^{1/2} + \frac{1}{2} \phi''(\vartheta_d) h_d ^ 2 \Dt \right)^4}{\Dt} \right] \right| = 0
\end{equation*}
with $\vartheta_d \in (-h_d \sqrt{\Dt}, h_d \sqrt{\Dt})$. Note that The term inside the expectation is composed of a sum of terms of the form $k h_d^n \phi''(\vartheta_d)^m \Dt^\alpha$ for integers $n,m \geq 0$ and reals $\alpha > 0$, $k \in \mathbb{R}$. This results from repeated applications of the Cauchy–Schwarz inequality and \cref{lem:bounds} as we did previously to prove (1).

Regarding (2), we can compute $\E[\Delta x (\Delta x)\tran]/\Dt$ instead of $\V[\Delta x]/\Dt$ as in the infinitesimal limit of $\Dt \downarrow 0$ the two quantities have to agree due to the convergence of the infinitesimal mean that we have already established. Hence by first order Taylor expansion of $\phi$ around 0 we need to show that for $d,u = 1, \dots, D$ and $R>0$:
\begin{equation*}
    \lim_{\Dt \downarrow 0} \sup_{\norm{x} < R} \Bigg| \sigma_x^2(x)_{d,u} - \E\left[ \frac{\big(\phi'(0) h_d \Dt^{1/2} + \frac{1}{2} \phi''(\vartheta_d) h_d ^ 2 \Dt \big) \big(\phi'(0) h_u \Dt^{1/2} + \frac{1}{2} \phi''(\vartheta_u) h_u ^ 2 \Dt \big)}{\Dt} \right] \Bigg| = 0
\end{equation*}
with $\vartheta_d \in (-h_d \sqrt{\Dt}, h_d \sqrt{\Dt})$, $\vartheta_u \in (-h_u \sqrt{\Dt}, h_u \sqrt{\Dt})$. The only term inside the expectation not vanishing in $\Dt$ is 
\begin{align*}
    \E[\phi'(0)^2 h_d h_u] &= \phi'(0)^2 \V[\varepsilon^W \psi(x) + \varepsilon^b]_{d,u}\\
    &+ \phi'(0)^2 \left(\mu^b_d + \mu^W_d \psi(x)\right)\left(\mu^b_u + \mu^W_u \psi(x)\right) \Dt\\
    &= \sigma_x^2(x)_{d,u} + \phi'(0)^2 \left(\mu^b_d + \mu^W_d \psi(x)\right)\left(\mu^b_u + \mu^W_u \psi(x)\right) \Dt.
\end{align*}
The (uniform on compacts) convergence of all terms aside from $\sigma_x^2(x)_{d,u}$ to $0$ once again follows from repeated applications of the Cauchy–Schwarz inequality and \cref{lem:bounds}.

Now, the continuity of $\mu_x(x)$ and $\sigma_x(x)$ are a consequence of the continuity of the conditional covariance $\V[\varepsilon^W \psi(x) + \varepsilon^b]$, and as $\V[\varepsilon^W \psi(x) + \varepsilon^b]$ is positive semi-definite so is $\sigma_x^2(x)$. Hence all the conditions of Assumption 2.1 hold true.

Finally, as $\psi$ is differentiable two times with continuity, it follows from the dependency of $\mu_x$ and $\sigma_x^2$ on $x$ only through $\V[\varepsilon^W \psi(x) + \varepsilon^b]$ that Assumption 2.2 is satisfied too. The application of Theorem 2.1 completes the proof.
\end{proof}

\begin{proof}[Proof of Corollary 3.1]
Notice that
\begin{equation*}
    d[W \psi(x)]_t + d[b]_t = d[W \psi(x) + b]_t = \diag(\V[\varepsilon^W_t \psi(x_t) + \varepsilon^b_t|x_t])dt
\end{equation*}
Then expanding $dW_t$ and $db_t$ in (12) shows that the drift terms are matched between (11) and (12). The quadratic variation of (11) is
\begin{equation*}
    \phi'(0)^2 \diag(\V[\varepsilon^W_t \psi(x_t) + \varepsilon^b_t|x_t]) dt
\end{equation*}
which is equal to the quadratic variation of (12) as it is computed as
\begin{equation*}
    d[x]_t = d[\phi'(0)(W \psi(x) + b)]_t = \phi'(0)^2 d[W \psi(x) + b]_t
\end{equation*}
This shows the equivalence in law between the solution of (11) and the solution of (12). Then (13) immediately follows by direct computation.
\end{proof}

\begin{proof}[Proof of Corollary 3.2 and Corollary 3.3]
Notice that
\begin{align*}
    &d[W\psi(x^{(i)}) + b, W\psi(x^{(j)}) + b]_t\\ 
    &\quad=\C[\varepsilon^W_t \psi(x_t^{(i)}) + \varepsilon^b_t, \varepsilon^W_t \psi(x_t^{(j)}) + \varepsilon^b_t|x_t^{(i)},x_t^{(j)}]dt\\
    &\quad=\big(\Sigma^b + \C[\varepsilon^W_t \psi(x_t^{(i)}), \varepsilon^W_t \psi(x_t^{(j)})|x_t^{(i)},x_t^{(j)}]\big)dt
\end{align*}
and
\begin{align*}
    &\C[\varepsilon^W_t \psi(x_t^{(i)}), \varepsilon^W_t \psi(x_t^{(j)})|x_t^{(i)},x_t^{(j)}]_{r,c}\\
    &\quad=\E[(\varepsilon^W_{t,r,\bullet} \psi(x_t^{(i)}))(\varepsilon^W_{t,c,\bullet} \psi(x_t^{(j)}))|x_t^{(i)},x_t^{(j)}]\\
    &\quad=\sum_{d,u=1}^{D} \psi(x_{t,d}^{(i)}) \psi(x_{t,u}^{(j)}) \E[W_{r,d} W_{c,u}]\\
    &\quad=\Sigma^{W_O}_{r,c} \sum_{d,u=1}^{D} \psi(x_{t,d}^{(i)}) \psi(x_{t,u}^{(j)}) \Sigma^{W_I}_{d,u}\\
    &\quad=\Sigma^{W_O}_{r,c} (\psi(x_t^{(i)})\tran \Sigma^{W_I} \psi(x_t^{(j)})).
\end{align*}
This proves Corollary 3.2. Corollary 3.3 follows by setting $\sigma^b = \sigma_b \I_D$, $\sigma^{W_I} = \I_D$ and $\sigma^{W_O} = \sigma_w D^{-1/2} \I_D$.
\end{proof}

\section{Additional experiments and plots}

\subsection{Bayesian inference}

\begin{figure}[!htb]
    \centering
    \includegraphics[width=\linewidth]{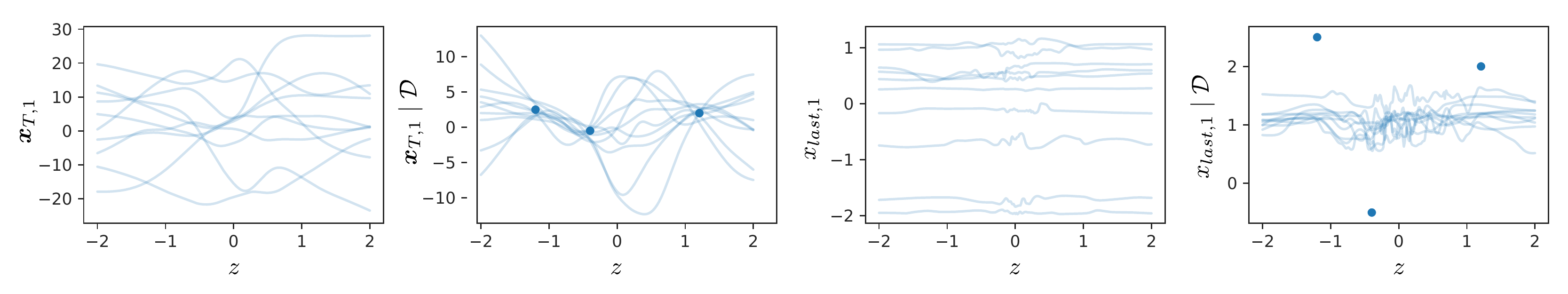}
    \caption{10 function samples in light blue over $z \in [-2,2]$ for: prior $\xdt_{T,1}$ and ABC-posterior $\xdt_{T,1} | \mathcal{D}$ for $\mathcal{SR}_{\tanh}$ (2 leftmost); prior $x_{last,1}$ and ABC-posterior $x_{last,1} | \mathcal{D}$ for $\mathcal{EO}_{\tanh}$ (2 rightmost); 3 data points of $\mathcal{D}$ in blue.}
    \label{fig:abc}
\end{figure}

 In this toy experiment we perform approximate Bayesian inference via Approximate Bayesian Computation (ABC, \citep{sisson2018handbook}) rejection sampling for function regression. We consider the setting of Assumption 3.4 with $\phi = \tanh$, $\sigma_w^2=\sigma_b^2=10$, $T=1$ and $L=D=500$. For this experiment we use a random input layer given by $\xdt_0 = W_I z$ where $W_I \in \R^{D \times 1}$ and $W_{I,d,1} \overset{i.i.d.}{\sim} \mathcal{N}(0,1)$ which makes the distribution of $\xdt_{T,1}$ symmetric around 0. We refer to this model as $\mathcal{SR}_{\tanh}$. We fix a computational budget of $10.000$ simulations and compute $10$ approximate posteriors samples by selecting the $10$ function draws with smallest $l_2$ distance from a synthetic dataset $\mathcal{D}$ consisting of $3$ data points. The results are reported in \cref{fig:abc} where we also compare with the results obtained by applying the same ABC algorithm to the first pre-activation of the last layer of $\mathcal{EO}_{\tanh}$.
We observe that the more flexible prior results in significantly improved efficiency with ABC: as the prior draws are almost constant in $\mathcal{EO}_{\tanh}$ we are constrained to finding the line with minimum distance from the 3 points. The use of $\sigma_w^2=\sigma_b^2=10$ in $\mathcal{SR}_{\tanh}$ compared to $\mathcal{SC}_{\tanh}$ allows for increased range and variability of $\xdt_{T,1}$. While it's possible to similarly increase weight and bias variances in $\mathcal{EO}_{\tanh}$ while remaining on the edge of chaos, this does not solve the underlying issue that the model a priori (hence a posteriori) assigns all probability mass to constant functions in the limit of $L \uparrow \infty$. Simulations (not shown) confirms that this modification does not improve the posterior inference efficiency for $\mathcal{EO}_{\tanh}$. It should be noted that we do not advocate the use of ABC rejection sampling as a realistic solution for this inference setting. Nonetheless this toy experiment exemplifies how a prior-data conflict typically frustrates inference algorithms.

\subsection{Additional experiment for Section 4.1}

\begin{figure}[!htb]
    \centering
    \includegraphics[width=0.75\linewidth]{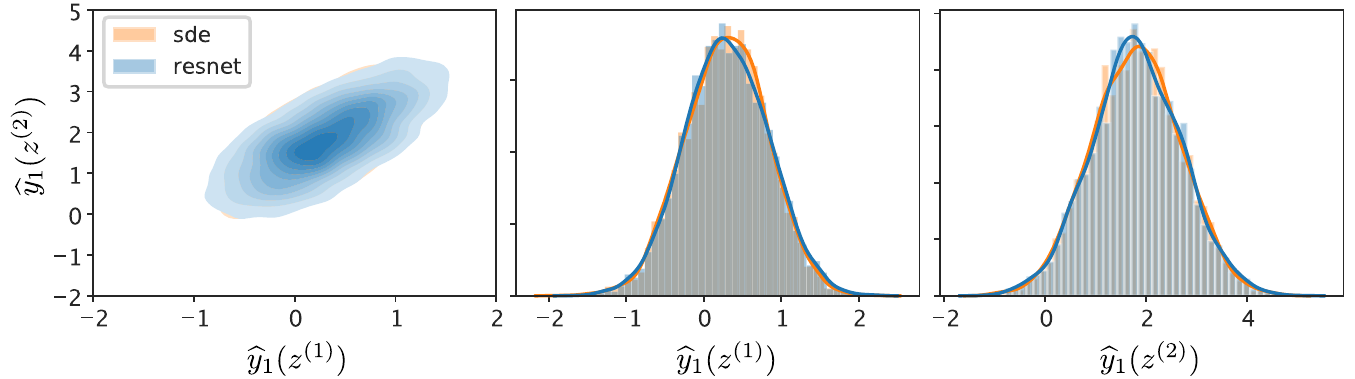}
    \caption{For model $\mathcal{SC}_{\swish}$: 2D KDE plot for $(\widehat{y}_1(z^{(1)}),\widehat{y}_1(z^{(2)}))$ (left), 1D KDE and histogram plots for $\widehat{y}_1(z^{(1)})$ (center), $\widehat{y}_1(z^{(2)})$ (right) when $\widehat{y}_1$ is sampled from a ResNet (resnet) and from the Euler discretization of its limiting SDE (sde); $\widehat{y}$ denotes a generic model output, hence $\widehat{y}_1$ is its first dimension.}
    \label{fig:f_swish}
\end{figure}

We replicate the experiment of Section 4.1 for the swish activation and plot the results in \cref{fig:f_swish} where again good agreement is observed.

\subsection{Additional plots for Section 4.2}

\begin{figure}[!htb]
    \centering
    \includegraphics[width=\linewidth]{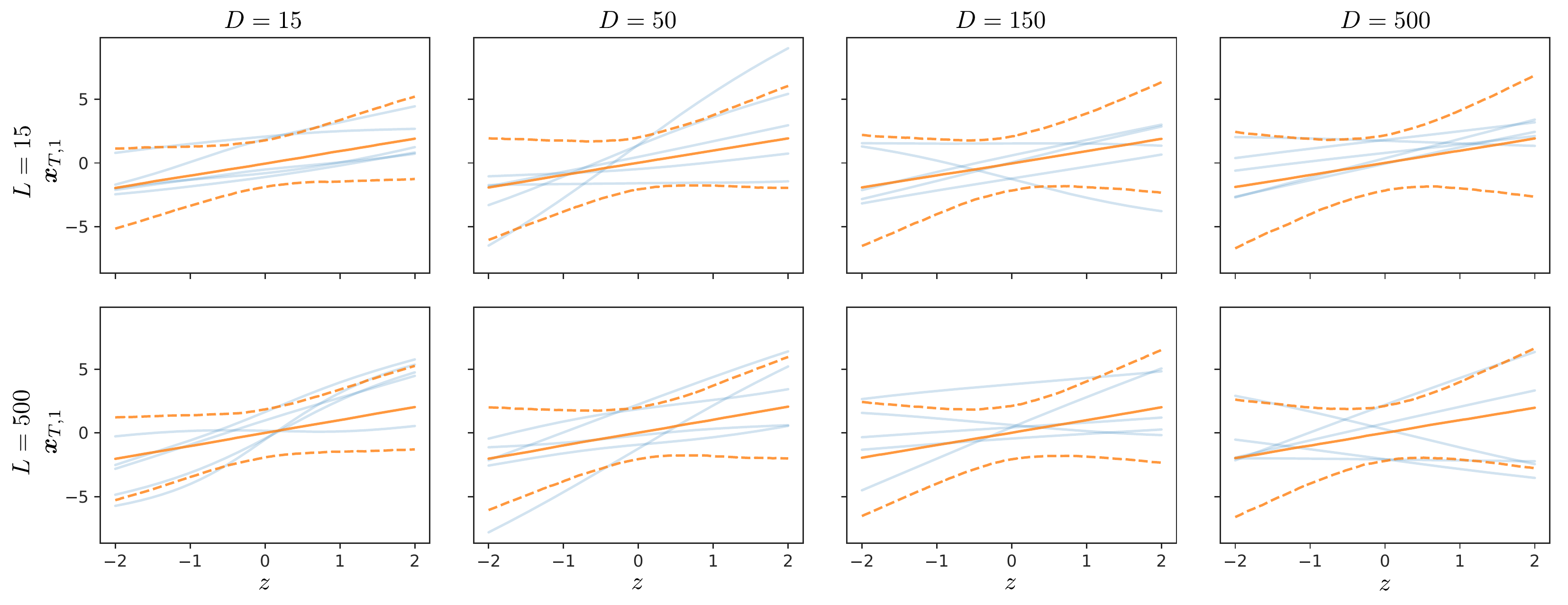}
    \includegraphics[width=\linewidth]{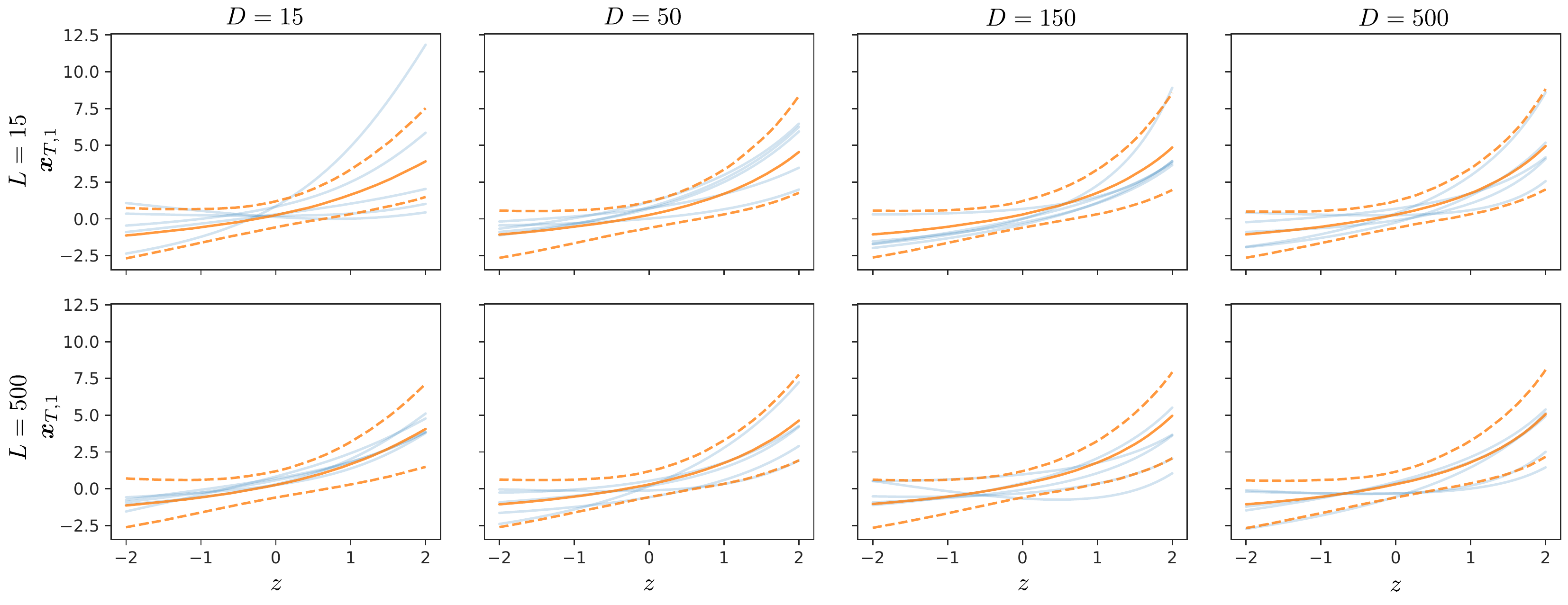}
    \caption{Function samples of $\xdt_{T,1}$ for $\mathcal{SC}_{\tanh}$ (top) and ${SC}_{\swish}$ (bottom) for different values of $L$ and $D$.}
    \label{fig:f_diffusion_ext}
\end{figure}

In \cref{fig:f_diffusion_ext} we plot additional function samples for the models $\mathcal{SC}_{\tanh}$ and ${SC}_{\swish}$ of Section 4.2 corresponding to different combinations of $L$ and $D$. We observe similar dynamics across different orders of magnitude for both $L$ and $D$.

\subsection{Additional 2D plots}

\begin{figure}[!htb]
    \centering
    \includegraphics[width=0.45\linewidth]{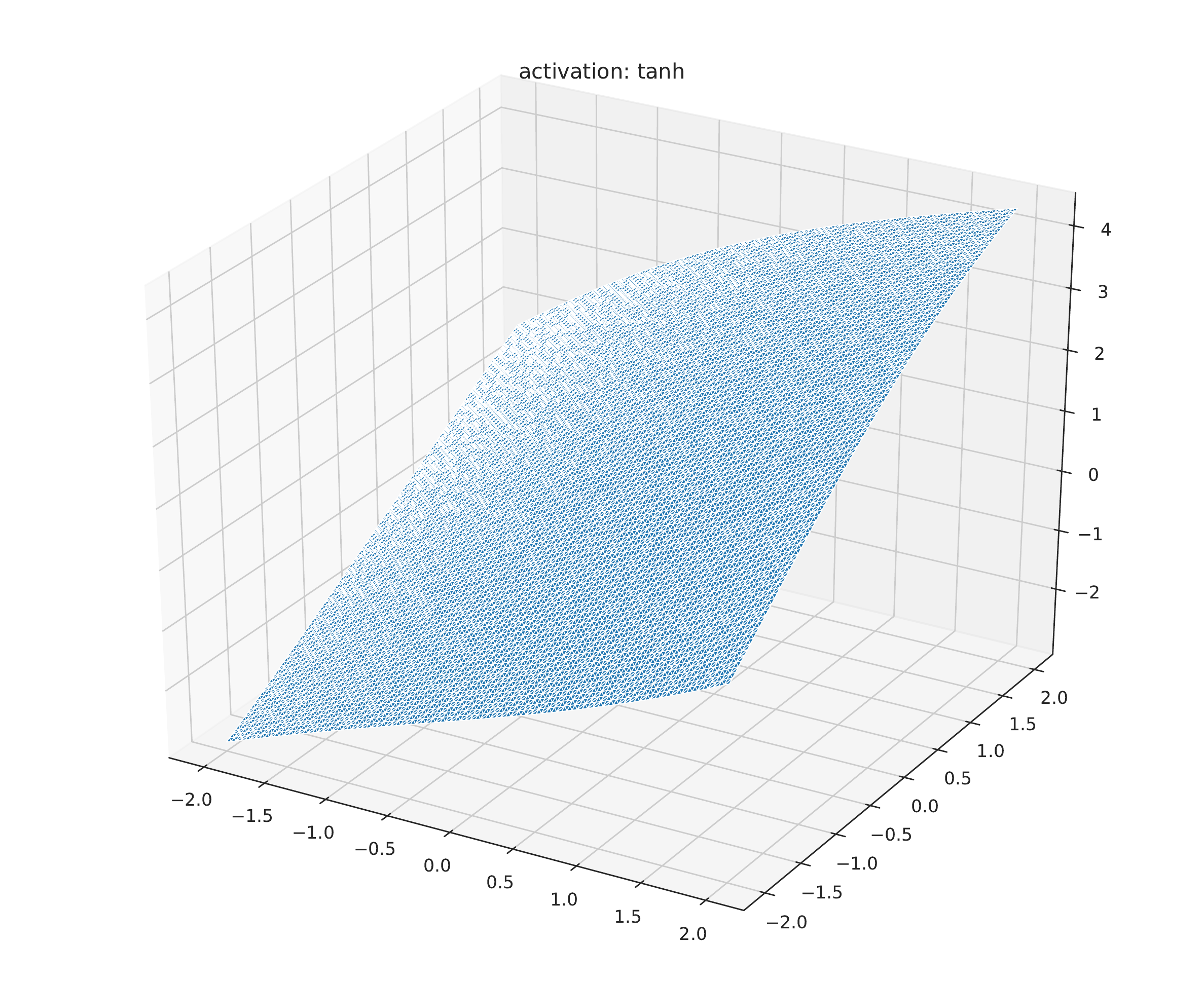}
    \includegraphics[width=0.45\linewidth]{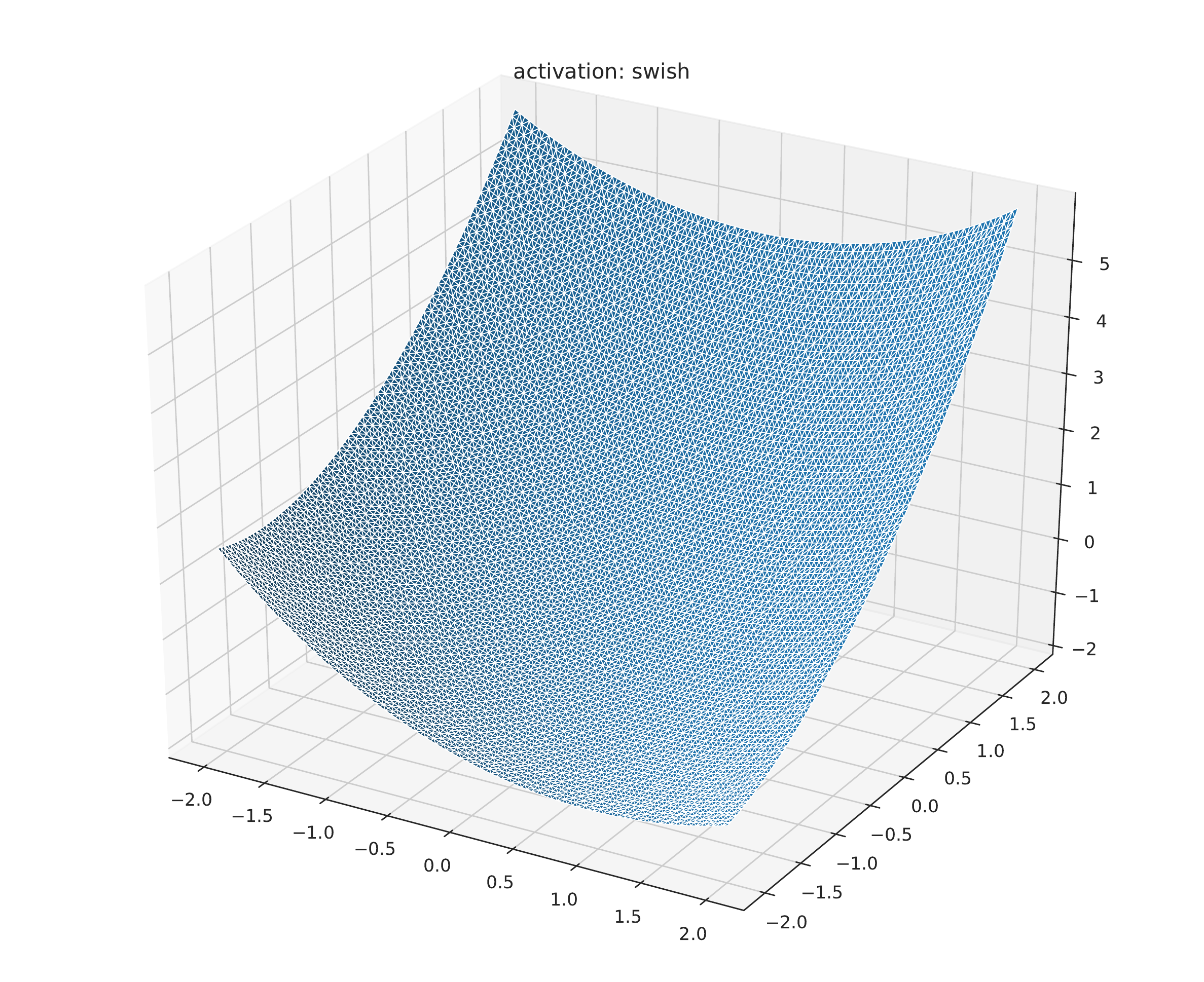}
    \caption{Function samples of $\xdt_{T,1}$ for $\mathcal{SC}_{\tanh}$ (left) and ${SC}_{\swish}$ (right) for $L=100$ and $D=100$ on the bounded interval $[-2,2] \times [-2,2]$.}
    \label{fig:f_2d}
\end{figure}

In \cref{fig:f_2d} we plot 2D function samples of $\xdt_{T,1}$ for $\mathcal{SC}_{\tanh}$ and ${SC}_{\swish}$ to complement the visualizations of Section 4.2.

\section{Related work}

In this section we discuss more in detail further connections with related work. \cite{chen2018neural} investigates the connection between infinite ResNets and ordinary differential equations (ODE), with a focus on potential computational advantages from gradient-descent training perspective. A difference between our work and that of \cite{chen2018neural} is that we only operate on the distribution of the model parameters, while \cite{chen2018neural} modifies the ResNet recursion with a $\Dt$ multiplicative term. The two approaches are equivalent for $\ReLU$ activations (see Section 3.4), where the SDE limit collapses to a deterministic ODE. While the focus of \cite{chen2018neural} is on the training of neural networks in the present work the emphasis is on the priori distribution of neural networks induced by a class of priors on the model parameters.

\cite{pennington2017resurrecting,pennington2018emergence} investigates the properties of the spectrum of the input-output Jacobian of deep neural networks at initialization which affects gradient-descent training speed. Of particular interest to the setting of the present work are the orthogonal initializations, proposed in \cite{pennington2017resurrecting} to achieve dynamical isometry, which seem amenable to a modification of the approach presented in our paper.

While \cite{pennington2017resurrecting,pennington2018emergence} relies on mean-field assumptions, i.e. infinite width and iid finite variance initializations, \cite{burkholz2019initialization} consider the setting of finite width and $\ReLU$ activations to derive exact results. Unfortunately, due to the use of $\ReLU$ activations the proposed initialization scheme is not of particular interest to our setting (see our discussion above regarding \cite{chen2018neural}).

\cite{hanin2018start,hanin2018neural} study ReLU networks when the width and depth are comparable and both large. As hinted in Section 4.2 and in the Discussion it would be interesting to investigate the behavior of the limiting SDEs when both depth and width grow unbounded. In this setting it is possible for the order of the limits and for the relative speed of convergence of $L$ and $D$ to affect the limiting dynamics.

Finally there has been recent interest in using heavy-tailed distributions for gradient noise and for trained parameter distributions, see for instance \cite{simsekli2019tail,martin2019traditional}. The present work covers exclusively Gaussian initializations and could be extended with some effort to cover finite-variance ones. Extensions to heavy tailed distributions would me more involved and likely resulting in less tractable Semimartingales (instead of SDEs) due to the presence of finite and infinite activity jump components. On the other hand the added flexibility could prove beneficial in bridging the gap between the performance of finitely trained neural networks and their limiting stochastic processes counterparts at least in some settings.

\bibliographystyle{apalike}
\bibliography{refs-new}